\documentclass[10pt,twocolumn]{style}

\title{Beyond Viewpoint Generalization: What Multi-View Demonstrations Offer and How to Synthesize Them for Robot Manipulation?}

\author{%
{\small
\textbf{Boyang Cai}\authmark{1,2,*}\quad
\textbf{Qiwei Liang}\authmark{1,2,*}\quad
\textbf{Jiawei Li}\authmark{2,*}\quad
\textbf{Shihang Weng}\authmark{2,*}\quad
\textbf{Zhaoxin Zhang}\authmark{2,*}
\textbf{Tao Lin}\authmark{3}\quad
\textbf{Zhengtu Liang}\authmark{2}\quad\\
\textbf{Xiangyu Chen}\authmark{1}\quad
\textbf{Wenjie Zhang}\authmark{1}\quad
\textbf{Jiaqi Mao}\authmark{4}\quad
\textbf{Weisheng Xu}\authmark{1}\quad
\textbf{Bin Yang}\authmark{1}\quad
\textbf{Jiaming Liang}\authmark{1,2}\quad
\textbf{Junhao Cai}\authmark{2}\quad
\textbf{Renjing Xu}\authmark{1,\textdagger}
}%
}

\affiliation{%
\authmark{1}The Hong Kong University of Science and Technology (Guangzhou)\\
\authmark{2}Shenzhen University
\authmark{3}Beijing Jiaotong University
\authmark{4}The Chinese University of Hong Kong, Shenzhen
}

\correspondence{%
\authmark{*}Equal contribution \quad
\authmark{\textdagger}Corresponding author
}

\date{}

\begin{document}
\maketitle

\begin{abstract}
Do multi-view demonstrations truly improve robot manipulation, or merely enhance cross-view robustness? We present a systematic study of the performance gains, scaling behavior, and underlying mechanisms of multi-view data for robot manipulation. To isolate the effect of viewpoint diversity, all policies are evaluated from a fixed canonical camera, so additional views are used only during training. Controlled experiments show that multi-view demonstrations consistently improve policy success across clean and visually randomized environments, multiple tasks, and different policy architectures. We further find that performance depends on view coverage: moderate viewpoint offsets provide effective supervision, whereas overly large offsets introduce less useful visual variation. Multi-view data also continues to improve overall policy performance after single-view scaling begins to saturate.
Mechanistic analysis suggests that these gains arise from better manipulation-centered representations rather than view invariance alone. Multi-view policies focus more on the gripper and manipulated objects, encode task-relevant pose in substantially richer nonlinear features, and yield a more reliable visual-to-action mapping. Motivated by both the value of multi-view supervision and its scarcity in large-scale robot datasets, we propose \textbf{RoboNVS}, a geometry-aware self-supervised framework that synthesizes novel-view videos from monocular demonstrations. RoboNVS constructs target-view geometric priors from metric-aligned, temporally stable depth and uses a video diffusion model with Bi-directional Masking to complete newly exposed regions. Compared with leading novel-view video methods, RoboNVS substantially reduces robot distortion, object disappearance or duplication. The synthesized views improve downstream policies in both simulation and real-world tasks, reaching 55--70\% success on three real-world Franka tasks compared with 5--25\% for monocular training.\enspace\href{https://youngyng.github.io/RoboNVS.github.io/}{\textbf{Project Page}}
\end{abstract}

\section{Introduction}
\label{sec:intro}

Robot manipulation policies are commonly trained from demonstrations collected under fixed and carefully controlled camera setups~\cite{chi2023diffusionpolicy,zhao2023learning,zhang2025flowpolicy,yan2025maniflow}.
During real-world deployment, however, camera viewpoints may differ from those seen during training, causing distribution shifts that can substantially degrade policy performance and generalization~\cite{xing2025shortcut,jiang2025you,liu2025geometry,li2026manivid,pang2025learning}.
A common solution is to collect multi-view demonstrations~\cite{yang2025novel,goyal2023rvt}, which expose the policy to the same manipulation process from different perspectives and reduce reliance on view-specific appearance cues.

Most existing explanations view multi-view data mainly as a tool for cross-view robustness and viewpoint invariance~\cite{jiang2025you,tian2025view}.
Under this view, its benefit should largely disappear when the policy is evaluated from the same camera used during training.
This raises a simple but important question:
\textbf{do multi-view demonstrations only improve robustness to camera changes, or can they also improve manipulation itself under fixed-view evaluation?}
Answering this question is important for robot data collection.
Scaling manipulation datasets usually requires collecting more demonstrations across diverse scenes, objects, and backgrounds~\cite{lin2024data,shi2025diversity}, which is costly and time-consuming.
In contrast, once a multi-camera setup is available, adding synchronized viewpoints to the same demonstration can provide extra supervision with relatively little additional collection effort.
If multi-view data improves policies even under single-view evaluation, it offers a practical and efficient way to scale robot learning datasets.

To study this question, we conduct a controlled empirical analysis that isolates the effect of viewpoint diversity.
We keep the policy architecture and training procedure fixed, vary only the viewpoint coverage and number of demonstrations, and evaluate all policies from a single canonical camera.
This ensures that any improvement cannot come from using extra views at test time.
We evaluate both clean and randomized environments, and analyze how performance changes with viewpoint offset, number of views, and demonstrations per view.
Our results reveal three main findings.
\textbf{First}, multi-view demonstrations improve manipulation success even when policies are evaluated from the original training viewpoint.
\textbf{Second}, moderate viewpoint offsets provide the strongest gains, while overly large view changes can hurt performance.
\textbf{Third}, multi-view supervision continues to improve performance in regimes where scaling single-view demonstrations begins to saturate.

We further analyze why multi-view data helps.
Spatial attribution shows that multi-view policies focus more on manipulation-relevant regions such as the end-effector and target objects.
Nonlinear probing shows that end-effector and object poses can be better decoded from multi-view visual representations, suggesting stronger nonlinear state encoding.
In addition, multi-view supervision improves the robustness of the visual-to-action mapping and leads to more stable policy optimization across training steps.

Despite these benefits, large-scale multi-view demonstrations remain scarce in existing robotic datasets~\cite{o2024open,wu2024robomind,khazatsky2024droid}.
Many datasets are monocular or collected from a small number of fixed cameras, and re-collecting synchronized multi-view demonstrations is often impractical.
To address this limitation, we propose \textbf{\hivepurple{RoboNVS}}, a geometry-aware self-supervised framework that synthesizes novel-view videos from monocular robot demonstrations.
RoboNVS first constructs target-view geometric priors from monocular videos, then uses a video diffusion model with Bi-directional Masking to complete missing regions.
The synthesized views serve as multi-view data augmentation and improve downstream imitation learning in both simulation and real-world tasks.

\textbf{Contributions.}
We make four contributions:
\keyterm{(1)} we show that multi-view demonstrations improve robot manipulation even under fixed single-view evaluation;
\keyterm{(2)} we analyze how viewpoint offsets, viewpoint coverage, and demonstration scale affect policy performance, identifying practical regimes where multi-view data is most effective;
\keyterm{(3)} we study the mechanisms behind these gains through spatial attribution, nonlinear probing, action-head robustness, and training dynamics; and
\keyterm{(4)} we introduce \textbf{RoboNVS}, a geometry-aware self-supervised framework that synthesizes multi-view demonstrations from monocular videos to scale viewpoint diversity for robot manipulation.

\begin{figure}[t]
  \centering
  \includegraphics[width=\linewidth]{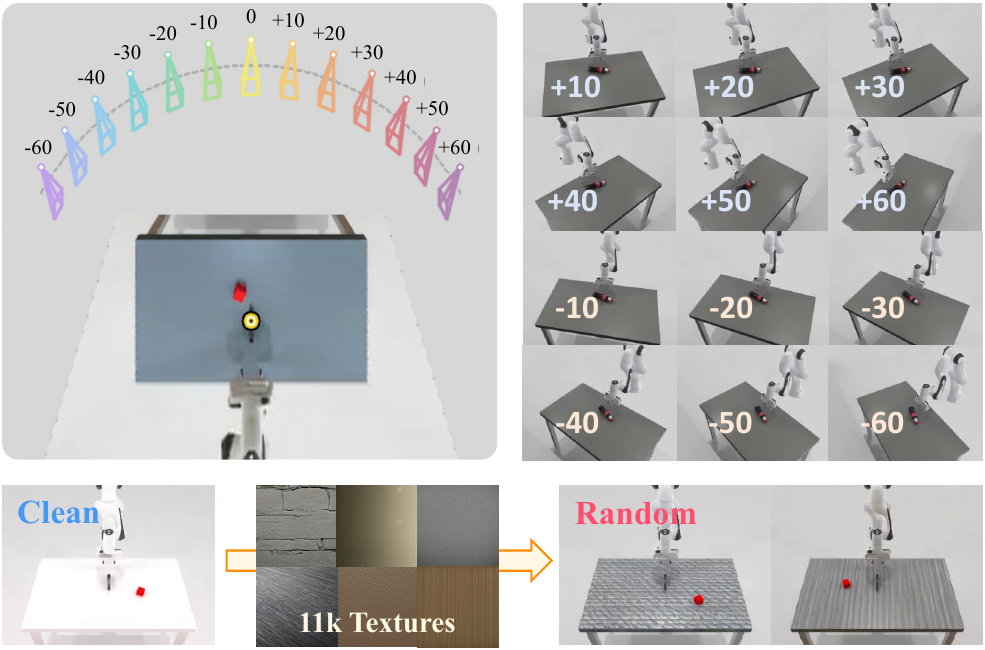}
  \caption{\textbf{Experimental setup overview.} Camera layout, example multi-view observations, and the Clean/Random environment modes.}
  \label{fig:setup_overview}
\end{figure}

\section{Multi-View Gains in Robot Learning}
\label{sec:understanding_mv}

Observing actions from multiple viewpoints has been shown to facilitate human skill acquisition in motor learning studies~\cite{wang2019multiple,rosendahl2024360,boucheix2018mixed}. This raises a natural question for robot learning: can robots also benefit from multi-view demonstrations, even when evaluated from a fixed viewpoint?
We first introduce the experimental setup (\cref{sec:setup}), then analyze the performance gains from multi-view training (\cref{sec:when_mv_help}), identify effective viewpoint ranges(\cref{sec:view_range}) and scaling behavior with demonstrations (\cref{sec:scaling_law}), and finally examine the mechanisms underlying these improvements (\cref{sec:mechanism}).

\subsection{Experimental Setup}
\label{sec:setup}

\noindent\textbf{Platform and tasks.}
Experiments are conducted in RoboTwin2.0~\cite{chen2025robotwin} with two embodiments: Franka for Diffusion Policy and Aloha for VLA.
We evaluate 12 manipulation tasks (AB, BBH, CA, CB, OL, MCP, MPA, SB, SBH, PCP, PS, PEC), reporting success rate over 50 evaluation rollouts. Further details of the task configurations are provided in Appendix Sec.~\ref{secC6}.

\noindent\textbf{Camera configuration.}
Cameras are placed on the upper hemisphere centered at the task workspace.
Each camera maintains a fixed distance of 1.70\,m and an elevation of $45^\circ$, forming a viewing cone.
Along the iso-elevation arc, azimuth angles are sampled every $10^\circ$ within $\pm60^\circ$, producing evenly spaced viewpoints of the workspace.

\noindent\textbf{Demonstrations.}
Unless otherwise specified, each task contains 50 demonstrations per viewpoint.
Single-view training uses demonstrations from the canonical $0^\circ$ camera, while multi-view training augments the dataset with synchronized viewpoints captured at additional azimuth angles.
For budget-matched experiments, we keep the number of demonstrations identical to the single-view setting by resampling from the multi-view dataset. The sampling is performed to preserve full coverage of the original single-view trajectories, ensuring that both settings observe the same set of manipulation scenarios while differing only in viewpoint supervision. Detailed sampling procedures are provided in Appendix~\ref{secCBudget-Matched}.

\noindent\textbf{Environment modes.}
We consider two environment settings.
\emph{Clean} uses a uniform white background, where variation mainly arises from object poses.
\emph{Random} samples tabletop textures from an 11k-texture pool at each rollout, introducing substantial appearance variation.

\noindent\textbf{Policy architectures.}
We evaluate two representative paradigms.
Diffusion Policy is trained using a ResNet-18 visual backbone and a three-frame observation window.
Vision-Language-Action (VLA) fine-tunes a pretrained $\pi_0$~\cite{black2024pi_0} model with LoRA~\cite{hu2022lora} for 5{,}000 steps on multi-task simulation data.

Figure~\ref{fig:setup_overview} provides an overview of the overall experimental setup and key components.
Additional details and training configurations are provided in the appendix.

\begin{figure*}[t]
    \centering
    \includegraphics[width=0.85\textwidth]{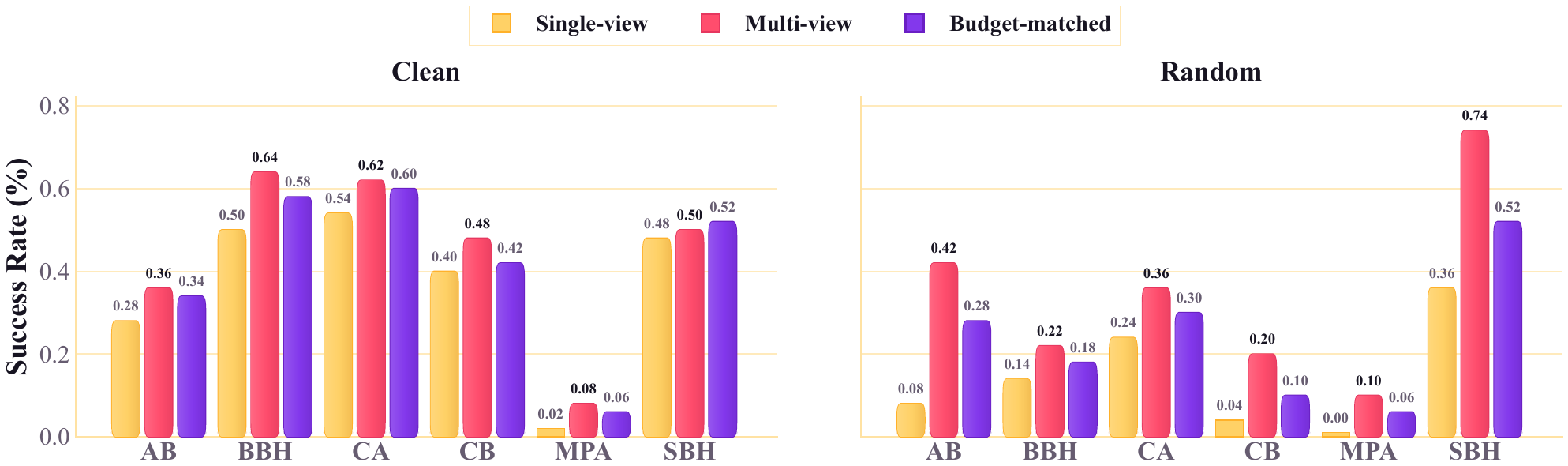}
    \caption{Comparison between single-view and multi-view training under clean and random settings.
    Multi-view training improves performance across tasks, with stronger gains under the random setting.}
    \label{fig:multi_vs_single}
\end{figure*}

\subsection{When Do Multi-View Demonstrations Help?}
\label{sec:when_mv_help}

Manipulation policies operate in both controlled workspaces and visually diverse environments.
To understand when multi-view supervision is most beneficial, we compare single-view and multi-view training under two modes: \textbf{Clean} and \textbf{Random}.
Single-view training uses demonstrations from the canonical $0^\circ$ camera, while multi-view training augments them with synchronized viewpoints at $\pm10^\circ$.
All policies are evaluated using only the $0^\circ$ camera to isolate the effect of multi-view supervision during training.
To decouple viewpoint diversity from data volume, budget-matched controls keep total videos and training steps fixed, assigning each trajectory to exactly one available viewpoint so only the viewpoint distribution changes.

\begin{table*}[t]
\centering
\caption{Multi-view supervision improves performance of a multi-task VLA model on both seen and unseen tasks.}
\label{tab:vla_multiview}
\begingroup
\small
\setlength{\tabcolsep}{3.6pt}
\renewcommand{\arraystretch}{1.16}
\resizebox{0.82\textwidth}{!}{%
\begin{tabular}{lccccccccc|ccc|ccc}
\toprule
\multirow{2}{*}{\textbf{Setting}}
& \multicolumn{9}{c|}{\textbf{Seen Tasks}}
& \multicolumn{3}{c|}{\textbf{Unseen Tasks}}
& \multicolumn{3}{c}{\textbf{Average}} \\
\cmidrule(lr){2-10}
\cmidrule(lr){11-13}
\cmidrule(lr){14-16}
& AB & BBH & CA & MCP & MPA & OL & PCP & PS & SBH
& SHB & CB & PEC
& Seen & Unseen & Overall \\
\midrule
Single-view
& 52 & 16 & \cellcolor{hiveHoney!20}\textbf{46} & 16 & 10 & 24 & 40 & 24 & 64
& 68 & 48 & 0
& 32.4 & 38.7 & 34.0 \\
\rowcolor{hiveSoftHoney!40}
Multi-view
& \textbf{68} & \textbf{28} & 40 & \textbf{30} & \textbf{14} & \textbf{28} & \textbf{52} & \textbf{34} & \textbf{72}
& \textbf{80} & \textbf{50} & \textbf{8}
& \textbf{40.7} {\footnotesize\textcolor{hiveMuted}{(+8.3)}}
& \textbf{46.0} {\footnotesize\textcolor{hiveMuted}{(+7.3)}}
& \textbf{42.0} {\footnotesize\textcolor{hiveMuted}{(+8.0)}} \\
\bottomrule
\end{tabular}%
}
\endgroup
\end{table*}

\begin{finding}[Multi-view training improves manipulation performance, especially under visual diversity.]\end{finding}

As shown in Fig.~\ref{fig:multi_vs_single}, multi-view supervision improves success rates even under single-view evaluation, indicating benefits beyond viewpoint mismatch.
The gains are larger in the Random setting, where stronger appearance variation increases reliance on view-specific cues and spurious background correlations during training.
Budget-matched comparisons further show that these gains cannot be explained by simply increasing the number of training videos.

\begin{finding}[The benefit transfers to pretrained VLA models and multi-task training.]\end{finding}
Table~\ref{tab:vla_multiview} reports results for a multi-task VLA model.
For \textbf{seen tasks}, the model is trained jointly on multiple tasks, while for \textbf{unseen tasks} it is fine-tuned and transferred to tasks not used during multi-task training.
Multi-view supervision improves the average success rate by +8.3 on seen tasks and +7.3 on unseen tasks, demonstrating that the advantage extends to large pretrained models and realistic multi-task pipelines across diverse manipulation task families.

\subsection{Which Viewpoint Range Works Best?}
\label{sec:view_range}

The previous results show that multi-view supervision improves manipulation performance even under fixed $0^\circ$ evaluation.
We next ask a more practical question: \textbf{which viewpoint range should be used?}
We isolate the effect of angular offset and study which range provides the most useful supervision.

We conduct two controlled experiments.
First, in the standard offset study, we train policies with the canonical view and a symmetric offset pair $\{0^\circ,-k^\circ,+k^\circ\}$, where $k\in\{10,20,30,40,50,60\}$.
This measures which angular offset gives the strongest policy performance when additional views are available.
Second, we perform a strict budget-matched VA experiment following the protocol introduced above.
For each $\pm k^\circ$ setting, the same trajectories are cycled through $-k^\circ$, $0^\circ$, and $+k^\circ$ so that only the viewpoint distribution changes.

\begin{table}[t]
\centering
\caption{
Task-wise success rates (\%) under each viewpoint offset in the standard offset study.
}
\label{tab:view_ablation_mean}
\begingroup
\small
\setlength{\tabcolsep}{3.8pt}
\renewcommand{\arraystretch}{1.10}
\resizebox{0.8\linewidth}{!}{%
\begin{tabular}{lccccccc}
\toprule
\textbf{Task}
& $0^\circ$
& $\pm10^\circ$
& \cellcolor{hiveHoney!12}$\pm20^\circ$
& $\pm30^\circ$
& $\pm40^\circ$
& $\pm50^\circ$
& $\pm60^\circ$
\\
\midrule
AB
& 8 & 42 & \cellcolor{hiveHoney!12}\textbf{80} & 70 & 64 & 56 & 48
\\
BBH
& 14 & 22 & \cellcolor{hiveHoney!12}\textbf{32} & 22 & 20 & 12 & 8
\\
CA
& 24 & 32 & \cellcolor{hiveHoney!12}\textbf{62} & 48 & 50 & 18 & 12
\\
CB
& 4 & 20 & \cellcolor{hiveHoney!12}\textbf{58} & 30 & 28 & 18 & 10
\\
\midrule
Mean
& 12.5 & 29.0 & \cellcolor{hiveHoney!12}\textbf{58.0} & 42.5 & 40.5 & 26.0 & 19.5
\\
\bottomrule
\end{tabular}%
}
\endgroup
\end{table}

\begin{finding}[Moderate viewpoint ranges are consistently effective across manipulation tasks.]\end{finding}
Table~\ref{tab:view_ablation_mean} shows that moderate viewpoint offsets consistently improve performance over the canonical-view baseline.
The strongest performance appears at $\pm20^\circ$, while larger offsets degrade performance.
This suggests that useful target views should provide complementary geometry while remaining visually compatible with the canonical manipulation view.

\begin{table}[t]
\centering
\caption{
Strict budget-matched VA experiment under fixed $0^\circ$ evaluation.
All budget-matched settings use the same data budget and training steps; the rightmost column reports the full three-view stack for reference.
}
\label{tab:budget_matched_va}
\begingroup
\small
\setlength{\tabcolsep}{3.8pt}
\renewcommand{\arraystretch}{1.10}
\resizebox{0.95\linewidth}{!}{%
\begin{tabular}{lcccccccc}
\toprule
\textbf{Task}
& $0^\circ$
& $\pm10^\circ$
& \cellcolor{hiveHoney!12}$\pm20^\circ$
& $\pm30^\circ$
& $\pm40^\circ$
& $\pm50^\circ$
& $\pm60^\circ$
& \shortstack{\textcolor{hiveMuted}{\textbf{$0^\circ+\pm10^\circ$}}\\\textcolor{hiveMuted}{Stack}} \\
\midrule
AB
& 8 & 50 & \cellcolor{hiveHoney!12}\textbf{58} & 48 & 44 & 44 & 42
& \textcolor{hiveMuted}{42} \\
BBH
& 14 & 18 & \cellcolor{hiveHoney!12}\textbf{22} & 14 & 12 & 12 & 8
& \textcolor{hiveMuted}{22} \\
CA
& 24 & 22 & \cellcolor{hiveHoney!12}\textbf{32} & 26 & 22 & 18 & 14
& \textcolor{hiveMuted}{32} \\
CB
& 4 & 14 & \cellcolor{hiveHoney!12}\textbf{22} & 16 & 14 & 10 & 8
& \textcolor{hiveMuted}{20} \\
\midrule
Mean
& 12.5 & 26.0 & \cellcolor{hiveHoney!12}\textbf{33.5} & 26.0 & 23.0 & 21.0 & 18.0
& \textcolor{hiveMuted}{29.0} \\
\bottomrule
\end{tabular}%
}
\endgroup
\end{table}

\begin{finding}[Multi-view gains are not merely driven by extra data volume under matched budgets.]\end{finding}
Table~\ref{tab:budget_matched_va} shows that moderate multi-view data still outperforms the $0^\circ$ baseline even when the total data budget and training steps are strictly matched.
In this setting, $\pm20^\circ$ achieves the best average success rate.
Since each trajectory appears exactly once, the improvement cannot be attributed to more data or longer training.
Instead, it comes from changing the viewpoint distribution itself.

Together, the two angle studies show that moderate viewpoint changes help more than overly large shifts, which may introduce appearance changes less aligned with manipulation geometry and harder for policies to reconcile.
In both the full and strict budget-matched settings, $\pm20^\circ$ performs best, suggesting effective geometric supervision without excessive visual variation. Additional stacked multi-view experiments are provided in Appendix~\ref{sec:stacked_multiview}.

\begin{figure*}[t]
\centering
\includegraphics[width=0.9\textwidth]{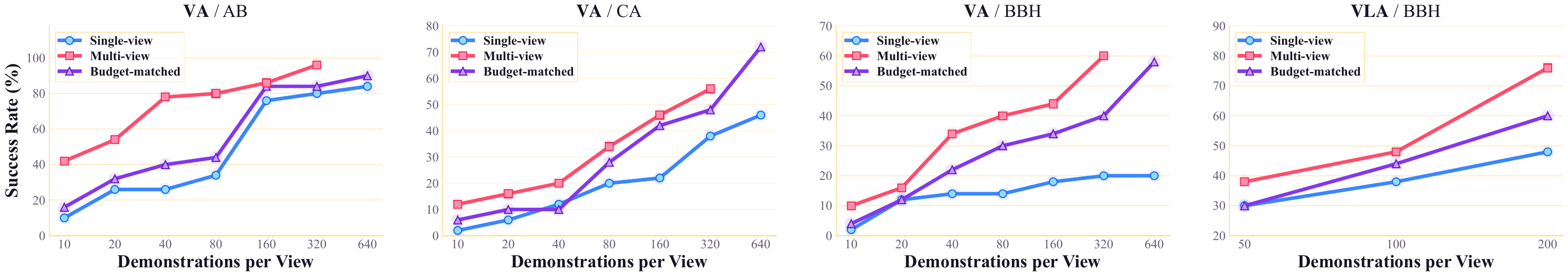}
\caption{
Scaling laws of single-view vs.\ multi-view training on manipulation tasks.
Multi-view data continues to improve performance when single-view training saturates, achieving comparable or higher performance with substantially less scene diversity.
}
\label{fig:scaling_law_plot}
\end{figure*}

\begin{figure}[t]
\centering
\includegraphics[width=0.9\linewidth]{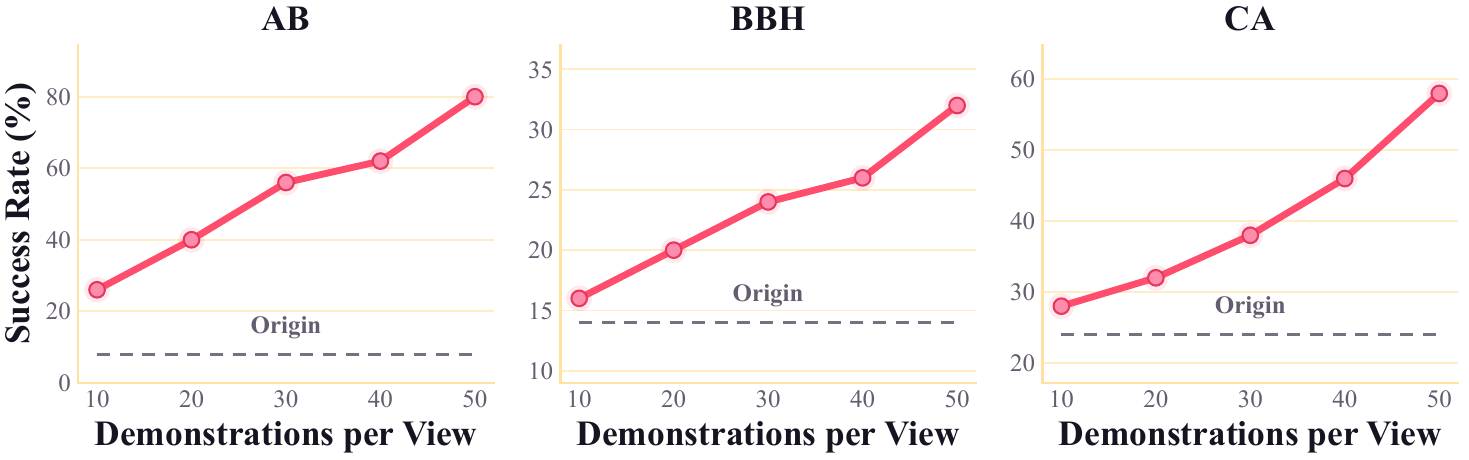}
\caption{
Scaling behavior of multi-view demonstrations under fixed $0^\circ$ evaluation.
Each subplot shows success rate versus demonstrations per added view ($10\!\rightarrow\!50$) for one task.
The black dashed line indicates the single-view baseline trained on the origin view.
}
\label{fig:demo_scaling}
\end{figure}

\subsection{Scaling Behavior of Demonstrations}
\label{sec:scaling_law}

After identifying effective viewpoint ranges, we study how performance scales with the number of demonstrations.
Unlike the previous subsection, which isolates angular range, this part focuses on data quantity.
All policies are still evaluated under the fixed $0^\circ$ camera. We perform two controlled experiments.
First, fixing each task to its effective viewpoint range, we increase the number of demonstrations per added view from \textbf{10} to \textbf{50} (Fig.~\ref{fig:demo_scaling}).
Second, we scale the overall data budget: single-view demonstrations grow from \textbf{10} to \textbf{640}, while multi-view training augments the canonical view with the selected effective view range and assigns the same number of demonstrations to each view (Fig.~\ref{fig:scaling_law_plot}).

\begin{finding}[Multi-view gains scale with additional demonstrations across tasks.]\end{finding} 
With the viewpoint range fixed, increasing demonstrations per added view from 10 to 50 steadily improves performance across evaluated tasks (Fig.~\ref{fig:demo_scaling}). This indicates that multi-view supervision continues to benefit from additional demonstrations and does not saturate immediately. 

\begin{finding}[Multi-view training breaks single-view saturation under scaling.]\end{finding}
Figure~\ref{fig:scaling_law_plot} shows that single-view performance plateaus as demonstrations increase, whereas multi-view training continues to improve.
Scaling single-view data often requires collecting demonstrations across new scenes, object placements, or backgrounds.
In contrast, synchronized or synthesized multi-view data expands the supervision signal around the same manipulation trajectory.

\subsection{Dissecting the Mechanisms}
\label{sec:mechanism}

Multi-view supervision improves manipulation performance even when evaluation uses the original camera view.
This raises a key question: \textbf{what changes inside policies?}
We analyze three aspects of learned policies during training and execution: the visual representation, the visual-to-action mapping, and the optimization dynamics.

\noindent\textbf{Representation quality.}
We first examine how multi-view supervision reshapes visual representations.
Multi-view training uses $0^\circ$ together with symmetric $\pm20^\circ$ views, while evaluation remains fixed at $0^\circ$ for fair comparison.
We analyze representation quality from two complementary perspectives: where the policy attends spatially, and how task-relevant states are encoded in the learned features.

\begin{figure}[t]
\centering
\includegraphics[width=0.9\linewidth]{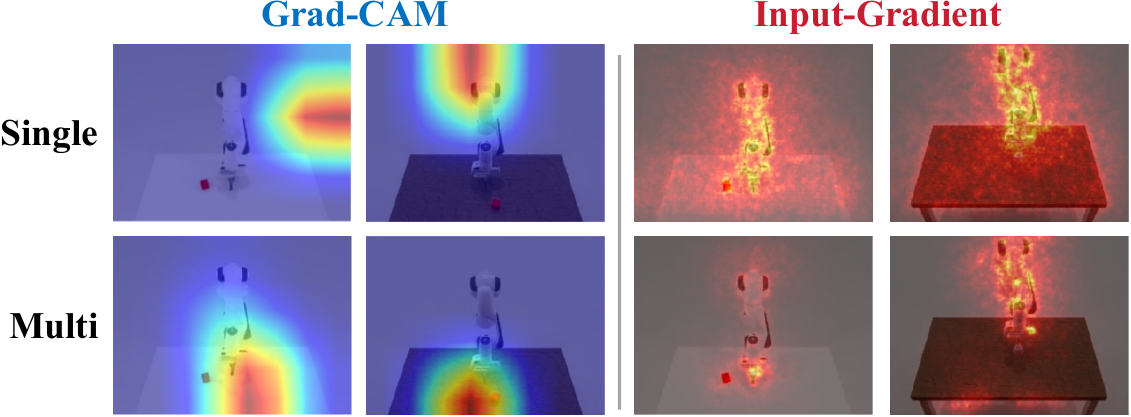}
\caption{
Grad-CAM and input-gradient analysis under fixed $0^\circ$ evaluation.
Multi-view supervision ($0^\circ+\pm20^\circ$) shifts attention toward the end-effector and manipulated objects, while single-view models exhibit stronger background sensitivity.
}
\label{fig:grad_analysis}
\end{figure}

\textit{Gradient attribution.}
To analyze spatial sensitivity, we compute Grad-CAM and input gradients with respect to predicted action magnitudes.
Specifically, we backpropagate the mean absolute pose change across all DoFs of the predicted action chunk to the visual encoder.
Grad-CAM highlights feature-level importance, while input gradients measure pixel-level sensitivity.
As shown in Fig.~\ref{fig:grad_analysis}, multi-view models concentrate more strongly on the end-effector and manipulated objects.
In contrast, single-view models show stronger sensitivity to background regions, suggesting a higher reliance on view-specific visual cues.

\begin{table*}[t]
\centering
\caption{
Clean-only probing results ($R^2$) for manipulation-relevant states under fixed $0^\circ$ evaluation.
Linear probing tests direct linear decodability ($\downarrow$), while nonlinear probing tests recoverability with a nonlinear decoder ($\uparrow$).
$\Delta$ denotes Multi-view $-$ Single-view.
}
\label{tab:linear_nonlinear_probe_clean_side_by_side}
\begingroup
\scriptsize
\setlength{\tabcolsep}{3.2pt}
\renewcommand{\arraystretch}{1.16}
\resizebox{0.9\textwidth}{!}{%
\begin{tabular}{lcc|cc|cc||cc|cc|cc}
\toprule
& \multicolumn{6}{c||}{\cellcolor{hiveHoney!18}\textbf{Linear Probe} ($R^2 \downarrow$)}
& \multicolumn{6}{c}{\cellcolor{hiveSoftMint}\textbf{Nonlinear Probe} ($R^2 \uparrow$)} \\
\cmidrule(lr){2-7}
\cmidrule(lr){8-13}
\textbf{Setting}
& \multicolumn{2}{c|}{\textbf{Adjust Bottle}}
& \multicolumn{2}{c|}{\textbf{Beat Block Hammer}}
& \multicolumn{2}{c||}{\textbf{Click Bell}}
& \multicolumn{2}{c|}{\textbf{Adjust Bottle}}
& \multicolumn{2}{c|}{\textbf{Beat Block Hammer}}
& \multicolumn{2}{c}{\textbf{Click Bell}} \\
\cmidrule(lr){2-3}
\cmidrule(lr){4-5}
\cmidrule(lr){6-7}
\cmidrule(lr){8-9}
\cmidrule(lr){10-11}
\cmidrule(lr){12-13}
& End-effector & Object
& End-effector & Object
& End-effector & Object
& End-effector & Object
& End-effector & Object
& End-effector & Object \\
\midrule

Single-view
& 0.510 & 0.502
& 0.816 & 0.678
& 0.930 & 0.952
& -0.112 & -0.032
& -1.220 & -0.777
& 0.283 & 0.529 \\

\textbf{Multi-view}
& \textbf{-0.116}
& \textbf{-0.116}
& \textbf{0.668}
& \textbf{0.540}
& \textbf{0.595}
& \textbf{0.736}
& \textbf{0.601}
& \textbf{0.166}
& \textbf{0.428}
& \textbf{0.511}
& \textbf{0.707}
& \textbf{0.835} \\

$\Delta$
& \textcolor{hiveCoral!95!black}{\textbf{-0.626}}
& \textcolor{hiveCoral!95!black}{\textbf{-0.618}}
& \textcolor{hiveCoral!95!black}{\textbf{-0.148}}
& \textcolor{hiveCoral!95!black}{\textbf{-0.138}}
& \textcolor{hiveCoral!95!black}{\textbf{-0.335}}
& \textcolor{hiveCoral!95!black}{\textbf{-0.216}}
& \textcolor{hiveMint!55!black}{\textbf{+0.713}}
& \textcolor{hiveMint!55!black}{\textbf{+0.198}}
& \textcolor{hiveMint!55!black}{\textbf{+1.648}}
& \textcolor{hiveMint!55!black}{\textbf{+1.288}}
& \textcolor{hiveMint!55!black}{\textbf{+0.424}}
& \textcolor{hiveMint!55!black}{\textbf{+0.306}} \\

\bottomrule
\end{tabular}%
}
\endgroup
\end{table*}

\textit{Linear and nonlinear probing.}
We further probe whether end-effector and object states can be decoded from frozen visual features.
For linear probing, we train linear regressors to predict manipulation-relevant states and report the coefficient of determination:
\begin{equation}
R^2 = 1 - \frac{\sum_i (y_i - \hat{y}_i)^2}{\sum_i (y_i - \bar{y})^2}.
\end{equation}
For nonlinear probing, we replace the linear regressor with a lightweight nonlinear decoder and evaluate whether the same states remain recoverable from the representation.

Table~\ref{tab:linear_nonlinear_probe_clean_side_by_side} reveals a more nuanced pattern.
Under linear probing, multi-view features yield lower $R^2$, indicating weaker direct linear decodability of end-effector and object poses.
However, under nonlinear probing, multi-view features consistently achieve higher $R^2$.
This suggests that multi-view training does not remove task-relevant state information.
Instead, it appears to reduce brittle linear pose shortcuts while preserving richer nonlinear structure that can better support downstream policy learning.

\noindent\textbf{Policy-head robustness.}
Representation changes alone do not fully explain policy improvement.
A policy must also map visual features to actions reliably.
To isolate this component, we freeze a pretrained DINOv2~\cite{oquab2023dinov2} encoder and train only the policy head.
Single-view training uses demonstrations from the canonical $0^\circ$ camera, while multi-view training combines $0^\circ$ with symmetric $\pm20^\circ$ views.
All evaluations are conducted under the fixed $0^\circ$ view.

\begin{table}[t]
\centering
\caption{Policy-head MSE loss under frozen DINOv2 features ($\downarrow$ lower is better).}
\label{tab:head_robustness}
\begingroup
\small
\setlength{\tabcolsep}{4pt}
\renewcommand{\arraystretch}{1.10}
\resizebox{0.95\linewidth}{!}{%
\begin{tabular}{lcc|cc|cc|cc}
\toprule
\multirow{3}{*}{\textbf{Metric}}
& \multicolumn{4}{c|}{\textbf{Random $\downarrow$}}
& \multicolumn{4}{c}{\textbf{Clean $\downarrow$}} \\
\cmidrule(lr){2-5}
\cmidrule(lr){6-9}
& \multicolumn{2}{c|}{\textbf{BBH}}
& \multicolumn{2}{c|}{\textbf{CA}}
& \multicolumn{2}{c|}{\textbf{BBH}}
& \multicolumn{2}{c}{\textbf{CA}} \\
\cmidrule(lr){2-3}
\cmidrule(lr){4-5}
\cmidrule(lr){6-7}
\cmidrule(lr){8-9}
& \textbf{Single} & \textbf{Multi}
& \textbf{Single} & \textbf{Multi}
& \textbf{Single} & \textbf{Multi}
& \textbf{Single} & \textbf{Multi} \\
\midrule
MSE
& 0.0128 & \textbf{0.0070}
& 0.0133 & \textbf{0.0087}
& 0.0104 & \textbf{0.0067}
& 0.0082 & \textbf{0.0074} \\
\bottomrule
\end{tabular}%
}
\endgroup
\end{table}

Table~\ref{tab:head_robustness} shows that multi-view supervision consistently reduces test MSE across both tasks and environment modes.
This indicates that multi-view training improves the robustness of the visual-to-action mapping, even when the visual encoder is fixed.
The effect is especially important under randomized backgrounds, where single-view policies are more likely to rely on spurious visual correlations.

\begin{figure}[t]
  \centering
  \includegraphics[width=\linewidth]{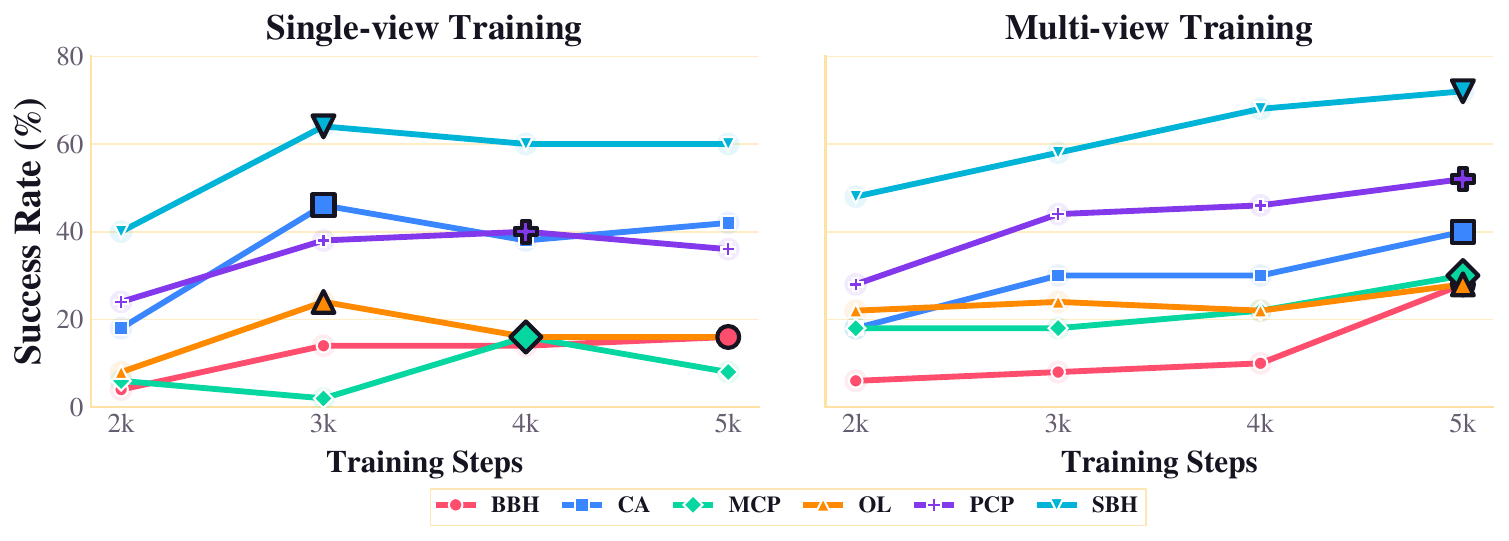}
  \caption{
  Training dynamics of a multi-task VLA model under single-view (left) and multi-view (right) supervision.
  Success rates are reported over training steps (2k--5k) across six tasks.
  Best checkpoints are highlighted.
  }
  \label{fig:optimization}
  \end{figure}

\noindent\textbf{Optimization dynamics.}
Finally, we analyze training trajectories.
Figure~\ref{fig:optimization} shows success rates over training steps for both settings.
Single-view training often exhibits unstable peaks at intermediate checkpoints, whereas multi-view supervision yields smoother improvements and higher final performance.
These results suggest that multi-view data improves optimization stability and sample efficiency, leading to policies that generalize more reliably.

\subsection{Multi-view versus 2D Augmentation}
\label{sec:mv_vs_2daug}

To test whether multi-view gains can be explained by generic image-level augmentation,
we compare against random perspective, random rotation, random cropping,
Gaussian blur, and a mixed augmentation setting under the same fixed data budget.
Each augmented dataset contains 50 samples, either 25 original $0^\circ$
demonstrations plus 25 augmented samples, or 10 originals plus four
augmentation types with 10 samples each. Sampling procedures and the
exact augmentation ranges are provided in Appendix~\ref{secCBudget-Matched}.

As shown in Table~\ref{tab:aug}, conventional 2D augmentations often
underperform the $0^\circ$ baseline, whereas view augmentation provides
gains under the same $0^\circ$ viewpoint. This suggests that the
benefit comes from viewpoint-dependent geometric supervision rather than
generic appearance perturbations.

\begin{table}[t]
\centering
\footnotesize
\caption{Budget-matched policy success rates of conventional 2D image augmentations versus multi-view augmentation, evaluated at the same fixed $0^\circ$ test viewpoint (SR\%).}
\label{tab:aug}
\begingroup
\setlength{\tabcolsep}{2.8pt}
\renewcommand{\arraystretch}{1.00}
\resizebox{\linewidth}{!}{%
\begin{tabular}{lcccc@{\hspace{0.85em}}lcccc}
\toprule
\textbf{Method} & \textbf{AB} & \textbf{BBH} & \textbf{CA} & \textbf{CB} &
\textbf{Method} & \textbf{AB} & \textbf{BBH} & \textbf{CA} & \textbf{CB} \\
\midrule
$0^\circ$
& 8 & 14 & 24 & 4
& Crop
& 4 & 8 & 16 & 2 \\
Perspective
& 6 & 2 & 6 & 2
& Blur
& 8 & 4 & 8 & 4 \\
Rotation
& 4 & 4 & 8 & 2
& Mixed
& 12 & 16 & 6 & 4 \\
\midrule
\rowcolor{hiveSoftMint!45}
\textbf{View Aug.\ ($\pm20^\circ$)}
& \textbf{34} & \textbf{20} & \textbf{30} & \textbf{18}
& & & & & \\
\bottomrule
\end{tabular}%
}
\endgroup
\end{table}

\begin{figure*}[t]
\centering
\includegraphics[width=0.9\textwidth]{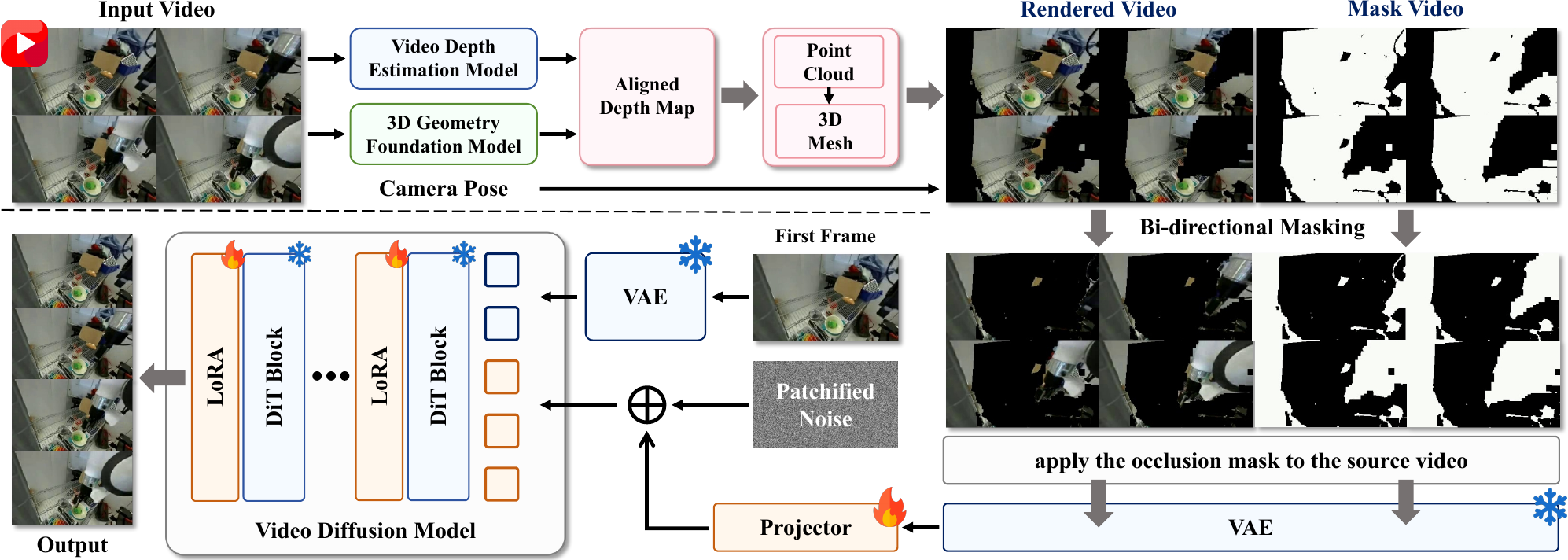}
\caption{
\textbf{Pipeline of RoboNVS.}
A monocular manipulation video is first converted into a temporally consistent and metric-aligned 3D representation.
The resulting DW-Mesh is rendered under target camera trajectories to produce RGB projections and occlusion masks.
These geometric priors condition a diffusion-based video completion model to synthesize physically plausible novel-view demonstrations.
}
\label{fig:pipeline}
\end{figure*}

\section{Scaling Multi-View Data with RoboNVS}
\label{sec:robonvs}

The previous sections reveal a simple but powerful message: \textbf{multi-view demonstrations are not just a robustness trick; they raise the learning ceiling of robot policies.} However, this also exposes a practical bottleneck at meaningful scale. Most large-scale robot datasets are still collected from one or a few fixed cameras, and re-collecting synchronized multi-view demonstrations is expensive, hardware-dependent, and often impossible for existing datasets.

This motivates a natural question:
\textbf{can we turn monocular robot videos into useful multi-view demonstrations?}
If so, multi-view supervision can be scaled without requiring new multi-camera data collection.
We therefore introduce \hivepurple{RoboNVS}, a geometry-aware novel-view synthesis framework that converts a monocular manipulation video into physically plausible target-view videos for policy learning.

Our goal is not merely to generate visually pleasing robot videos.
Instead, the generator serves as a \textbf{data-scaling engine}: it should synthesize additional viewpoints that preserve scene geometry, robot motion, and task-relevant spatial relations, so that the generated videos can be used as effective demonstrations for downstream imitation learning.

\subsection{Problem Setup}
\label{sec:nvs_problem}

Given a monocular manipulation video $V=\{I_t\}_{t=1}^{T}$ from a fixed camera, our goal is to synthesize a target-view video $\hat{V}=\{\hat{I}_t\}_{t=1}^{T}$ under a target camera trajectory $\{G_t\}_{t=1}^{T}$. The sequence should satisfy two requirements: \textbf{geometric consistency}, so that objects, robot links, and spatial layouts remain physically plausible; and \textbf{temporal consistency}, so that robot motion remains smooth across frames.

This problem is challenging for robotic videos.
Direct end-to-end camera-controlled video generation often lacks explicit 3D constraints, leading to distorted robot arms, duplicated objects, or hallucinated structures~\cite{bai2025recammaster,van2024generative}.
These artifacts are especially harmful for robot learning, because a policy may treat generated geometric errors as valid supervision.
RoboNVS therefore adopts a geometry-first design: we first render target-view geometric priors, and then use a diffusion model only to complete the missing regions.

\subsection{Overview}
\label{sec:nvs_overview}

Figure~\ref{fig:pipeline} illustrates the RoboNVS pipeline.
Given an input video, RoboNVS first estimates temporally consistent and metric-aligned depth maps, then converts them into a DW-Mesh representation~\cite{hu2025ex}.
Rendering the mesh under a target camera produces partial RGB projections and occlusion masks.
These geometry-aware signals condition a diffusion-based video completion model, which fills missing regions without hallucinating the entire scene from scratch.
This design is well suited for robot data augmentation, where geometric fidelity is more important than unconstrained visual creativity.
Details of self-supervised pair construction are provided in Appendix Sec.~\ref{secC3}.

\subsection{Depth Alignment}
\label{sec:depth_alignment}

Reliable novel-view synthesis requires accurate and temporally stable depth.
DepthCrafter~\cite{hu2025depthcrafter} provides temporally consistent relative depth with fine details, while DA3~\cite{lin2025depth} provides camera poses and metric depth.
We combine their strengths by aligning DepthCrafter depth to DA3 depth using a global scale-shift transformation in inverse depth space:
\begin{equation}
\alpha , \beta =
\arg\min_{\alpha , \beta}
\sum_{t}
\left\|
d_t^{-1} -
\left(
\alpha \tilde{d}_t^{-1} + \beta
\right)
\right\|_2^2 ,
\end{equation}
where $\tilde{d}_t$ is the relative depth from DepthCrafter and $d_t$ is the metric depth from DA3.
The aligned depth maps retain temporal smoothness while matching metric camera geometry.
We then convert them into DW-Mesh, which provides geometric projections and masks for video completion. We provide detailed comparisons of different depth prediction models in Appendix Sec.~\ref{secG2}.

\subsection{Video Completion}
\label{sec:diffusion_completion}

Rendering DW-Mesh under a target viewpoint gives a partial observation of the scene. Newly exposed regions, occluded robot parts, and background boundaries must still be completed in each frame. We formulate this as geometry-guided video inpainting.

Let $V=\{I_t\}$ be the input video and $M=\{M_t\}$ be the rendered occlusion masks.
We construct masked frames by
\begin{equation}
\tilde{I}_t = I_t \odot (1-M_t),
\end{equation}
where $\odot$ denotes element-wise multiplication.
The masked video $\tilde V$, mask sequence $M$, and reference frame $I_1$ are encoded using a frozen Wan2.1~\cite{wan2025wan} VAE and projected into the diffusion-transformer input space.
The first frame serves as an appearance reference to preserve object colors, robot textures, and scene identity.

We fine-tune Wan2.1-14B with LoRA adapters.
At diffusion time $\tau$, the model predicts the velocity field $v_\theta(z_\tau,c,\tau)$ conditioned on
\begin{equation}
c = \{I_1,\tilde V,M,p\},
\end{equation}
where $p$ is a fixed text prompt used for conditioning. The flow-matching objective is
\begin{equation}
\mathcal{L} =
\mathbb{E}_{\tau,z_\tau}
\left[
\|v_\theta(z_\tau,c,\tau)-v^\ast(z_\tau,\tau)\|_2^2
\right],
\end{equation}
where $v^\ast$ is the target velocity field.
This allows RoboNVS to complete missing regions while remaining anchored to rendered geometry, reducing physically implausible robot hallucinations.

\subsection{Bi-directional Masking}
\label{sec:bidirectional_masking}

A key challenge in self-supervised training is mask mismatch.
Standard inpainting masks often cover foreground objects, whereas novel-view rendering exposes peripheral regions, object boundaries, and newly visible backgrounds.
This mismatch can lead to blurred completions, ghosting, or hallucinated robot textures.

To reduce this gap, RoboNVS trains with both the rendered mask and its complement:
\begin{equation}
\bar{M}_t = 1 - M_t .
\end{equation}
The original mask encourages reconstruction of occluded robot and object regions, while the complementary mask forces completion of surrounding background and peripheral areas.
This increases mask diversity and improves robustness to viewpoint-induced occlusions.

For robot novel-view synthesis, the difficulty is not only what to inpaint, but also where missing regions appear.
Bi-directional masking reduces this mask distribution gap.

\subsection{Inference}
\label{sec:robonvs_inference}

At inference time, RoboNVS uses the same geometry pipeline without mask inversion.
Given a monocular video, we estimate aligned depth, build DW-Mesh, and render RGB projections and masks under the target camera.
The diffusion model then completes the masked regions using the source reference frame.
The output is a geometrically consistent novel-view video used as additional policy-training data, closing the loop between multi-view gains and monocular dataset scaling.

\section{RoboNVS Evaluation}
\label{sec:robonvs_experiments}

We evaluate RoboNVS along two axes.
First, we measure whether the generated videos are visually faithful and temporally consistent.
Second, we test whether the generated views improve downstream robot policies.
For robot learning, a useful generator must provide \textbf{policy-effective supervision}, not just high visual quality. We provide qualitative results of view synthesis and downstream policy training performance on RoboTwin in the Appendix Sec.~\ref{secF}.

\subsection{Implementation}
\label{subsec:implementation}

\noindent\textbf{Datasets.}
We train RoboNVS on approximately 10{,}000 manipulation videos from multiple public datasets.
The corpus primarily builds on Open X-Embodiment (OXE)~\cite{o2024open}, including Droid~\cite{khazatsky2024droid} and RT-1~\cite{brohan2022rt}.
We also include RH20T~\cite{fang2023rh20t}, RoboTwin~\cite{chen2025robotwin}, and RoboCoin~\cite{wu2025robocoin}, covering diverse embodiments, scenes, and manipulation behaviors.
Detailed statistics are provided in the Appendix Sec.~\ref{secC1}.

\noindent\textbf{Training.} We fine-tune Wan2.1-14B with LoRA rank $r=16$. Videos are resized to $320 \times 192$ with 49 frames. We use AdamW with learning rate $2\times10^{-5}$ for stable convergence. Training runs on 8 NVIDIA H200 GPUs with batch size 1 per GPU, giving an effective batch size of 8. The full fine-tuning process takes about one week.

\noindent\textbf{Metrics.} We report SSIM, LPIPS~\cite{zhang2018unreasonable}, and FVD~\cite{heusel2018ganstrainedtimescaleupdate} for spatial fidelity, perceptual quality, and temporal consistency. We also use VBench~\cite{huang2024vbench} to measure broader video quality and generation stability.

\noindent\textbf{Baselines.}
We compare against representative novel-view video generation methods.
One-stage camera-controlled models include ReCamMaster~\cite{bai2025recammaster} and ZeroNVS~\cite{sargent2024zeronvs}, where ZeroNVS is fine-tuned on Droid following VISTA~\cite{tian2025view}.
Two-stage inpainting-based methods include EX-4D~\cite{hu2025ex}, TrajectoryCrafter~\cite{yu2025trajectorycrafter}, and CogNVS~\cite{chen2025reconstruct}.
We also evaluate RoboNVS without Bi-directional Masking.

\noindent\textbf{Inference.} We use a 25-step denoising schedule. On a single NVIDIA RTX 4090 GPU with 48GB memory, RoboNVS generates a 49-frame video at $320\times192$ resolution in about 50 seconds per generated clip.

\subsection{Generation Quality}
\label{sec:generation_quality}

\begin{table}[t]
\centering
\caption{
Visual quality comparison of RoboNVS and baseline models.
$\uparrow$ indicates higher is better and $\downarrow$ indicates lower is better.
}
\label{tab:metrics}
\begingroup
\small
\setlength{\tabcolsep}{5pt}
\renewcommand{\arraystretch}{1.16}
\resizebox{0.9\columnwidth}{!}{%
\begin{tabular}{lcccc}
\toprule
\textbf{Method}
& \textbf{SSIM $\uparrow$}
& \textbf{LPIPS $\downarrow$}
& \textbf{FVD $\downarrow$}
& \textbf{VBench $\uparrow$} \\
\midrule
ZeroNVS-ft        & 0.3305 & 0.527 & 64.59 & 0.578 \\
ReCamMaster       & 0.3865 & 0.451 & 68.16 & 0.585 \\
CogNVS            & 0.4029 & 0.502 & 76.87 & 0.593 \\
TrajectoryCrafter & 0.3849 & 0.411 & 48.42 & 0.579 \\
EX-4D             & 0.4013 & 0.406 & 42.66 & 0.588 \\
\midrule
RoboNVS w/o Bi-Mask
                  & 0.4272 & 0.369 & 35.90 & 0.593 \\
\rowcolor{hiveSoftMint!45}
\textbf{RoboNVS}
                  & \textbf{0.4588} & \textbf{0.337} & \textbf{31.40} & \textbf{0.597} \\
\bottomrule
\end{tabular}%
}
\endgroup
\end{table}

We evaluate all methods on a held-out test set of 50 videos sampled from Droid, BridgeData V2~\cite{walke2023bridgedata}, and RoboCoin, ensuring that none of the sequences appear in training.
Table~\ref{tab:metrics} reports the quantitative results.
RoboNVS achieves the best performance across all metrics, improving pixel-level reconstruction, perceptual alignment, and temporal consistency.

The gain is most pronounced in FVD, suggesting that RoboNVS better preserves coherent robot motion across frames over long horizons. This is important for policy learning, because temporally inconsistent videos may provide misleading action-observation supervision even when individual frames look plausible.

\begin{figure*}[t]
\centering
\includegraphics[width=0.8\textwidth]{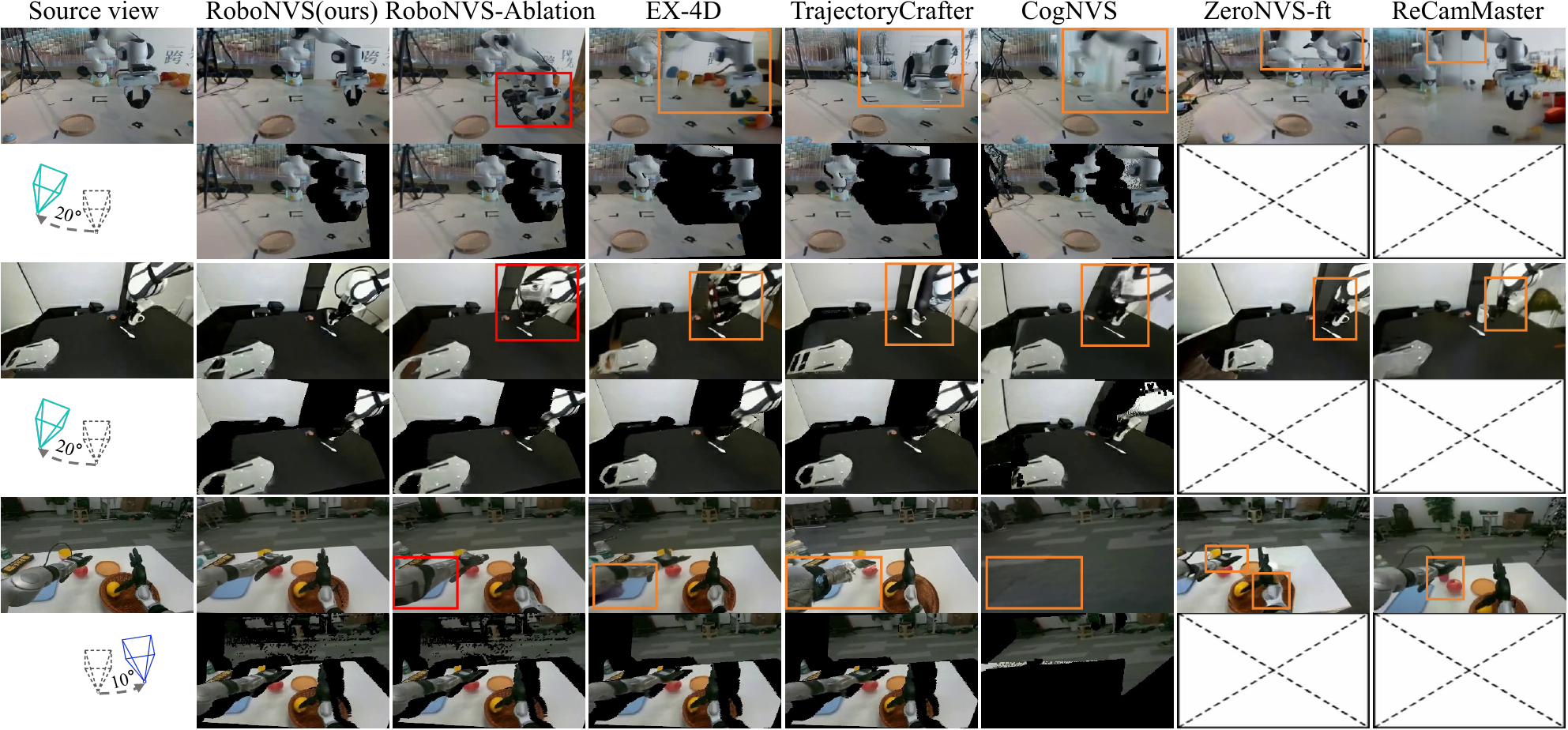}
\caption{
\textbf{Qualitative comparison with generative baselines.}
RoboNVS produces more geometrically consistent and temporally stable novel-view videos.
Baseline methods often exhibit object duplication, distorted robot structures, or incomplete mask filling.
}
\label{fig:fig_base_compare_new_realworld_visual_compare}
\end{figure*}

Figure~\ref{fig:fig_base_compare_new_realworld_visual_compare} provides qualitative comparisons.
Inpainting-based methods such as EX-4D, TrajectoryCrafter, and CogNVS often suffer from geometric distortions caused by inaccurate depth or incomplete robot reconstruction.
Camera-guided end-to-end models such as ReCamMaster may hallucinate duplicated objects or inconsistent structures.
In contrast, RoboNVS preserves the spatial layout of the scene and reconstructs robot structures more faithfully.
The comparison with RoboNVS w/o Bi-Mask further shows that Bi-directional Masking reduces ghosting and structural artifacts caused by mask distribution shift.

\subsection{Real-World Policy Evaluation}
\label{sec:real_world_eval}

Generation quality alone is not sufficient. A robot data generator must ultimately improve downstream policies. We therefore evaluate RoboNVS on real-world manipulation tasks on physical hardware using a 7-DoF Franka Panda with a top-down RealSense D435 camera.

\begin{figure}[t]
\centering
\includegraphics[width=\linewidth]{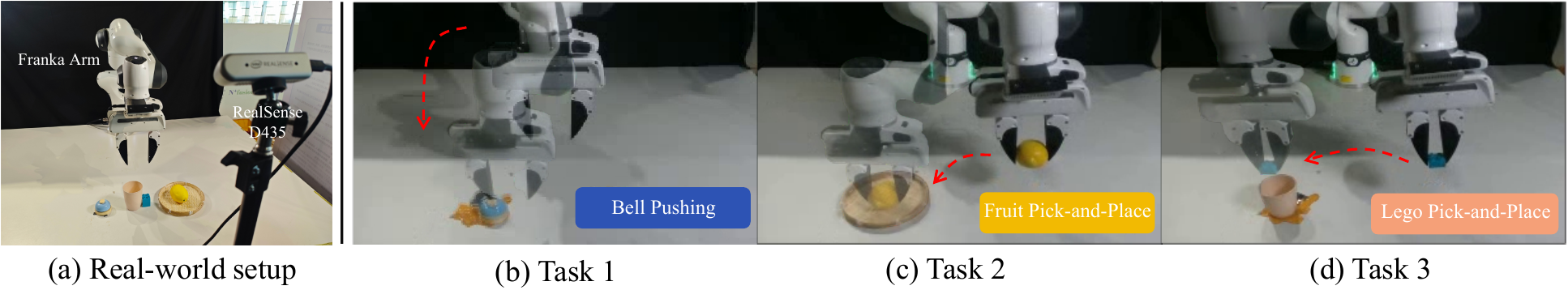}
\caption{
\textbf{Real-world setup and manipulation tasks.}
We evaluate RoboNVS on Bell Pushing, Fruit Pick-and-Place, and Lego Pick-and-Place using a Franka Panda robot.
}
\label{fig:task_demo}
\end{figure}

\noindent\textbf{Setup.}
We consider three representative tasks: \textit{Bell Pushing}, \textit{Fruit Pick-and-Place}, and \textit{Lego Pick-and-Place}.
Each task contains 30 expert trajectories collected from the base $0^\circ$ camera.
All policies are evaluated under the same $0^\circ$ viewpoint, with randomized object positions.
Success rates are reported over 20 independent physical robot trials.

\noindent\textbf{Augmentation protocols.}
We compare three synthetic multi-view augmentation methods:
\textbf{EX-4D}, \textbf{EX-4D w/ Depth Alignment} (EX-4D w/ DA), and \textbf{RoboNVS}.
Each method generates two target views:
$\{-20^\circ, 20^\circ\}$.
The generated videos are combined with the original monocular demonstrations to train a Diffusion Policy.

\begin{figure}[t]
\centering
\includegraphics[width=0.9\linewidth]{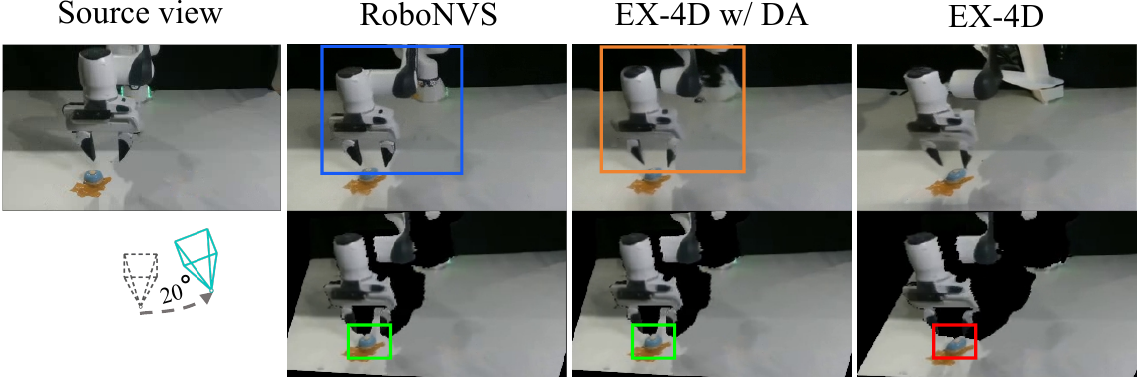}
\caption{
\textbf{Qualitative comparison on real-world tasks.}
Depth alignment improves target-view geometric masks, while RoboNVS further improves video completion quality, reconstructing robot arms and end-effectors more completely than EX-4D variants.
}
\label{fig:real_world_exp_compare}
\end{figure}

\noindent\textbf{Qualitative observations.}
Figure~\ref{fig:real_world_exp_compare} shows that EX-4D often produces distorted target-view geometry, especially around objects and robot structures.
Adding depth alignment improves the projected masks, but the completed videos still contain missing or blurry robot parts.
RoboNVS produces more complete mask filling and cleaner robot geometry, confirming the importance of both metric-temporal depth alignment and Bi-directional Masking.

\begin{table}[t]
\centering
\caption{
Policy success rates on real-world tasks.
Each result is evaluated over 20 trials.
}
\label{tab:success_rates}
\begingroup
\small
\setlength{\tabcolsep}{5pt}
\renewcommand{\arraystretch}{1.16}
\resizebox{0.9\columnwidth}{!}{%
\begin{tabular}{lccc}
\toprule
\textbf{Method}
& \textbf{Bell Pushing}
& \textbf{Fruit Pick-and-Place}
& \textbf{Lego Pick-and-Place} \\
\midrule
Monocular only
& 5/20 (25\%) & 2/20 (10\%) & 1/20 (5\%) \\
EX-4D
& 8/20 (40\%) & 6/20 (30\%) & 4/20 (20\%) \\
EX-4D w/ DA
& 10/20 (50\%) & 8/20 (40\%) & 7/20 (35\%) \\
\midrule
\rowcolor{hiveSoftMint!45}
\textbf{RoboNVS}
& \textbf{14/20 (70\%)}
& \textbf{12/20 (60\%)}
& \textbf{11/20 (55\%)} \\
\bottomrule
\end{tabular}%
}
\endgroup
\end{table}

\noindent\textbf{Quantitative analysis.}
As shown in Table~\ref{tab:success_rates}, RoboNVS-augmented policies outperform all baselines across the three real-world tasks.
The gain is especially clear on the most difficult Lego Pick-and-Place task, where RoboNVS reaches 55\% success, compared with 25\% for EX-4D and 5\% for the monocular baseline.

These results support the central motivation of RoboNVS. The generated videos are not only visually better; they provide useful geometric and spatial supervision for policy learning. Together with the multi-view analysis in Sec.~\ref{sec:understanding_mv}, this demonstrates a complete data-scaling loop: \textbf{multi-view demonstrations improve robot policies, and RoboNVS makes multi-view supervision available from monocular data at practical scale.}

\section{Conclusion}

We show that multi-view supervision can directly improve robot manipulation capability, beyond enhancing viewpoint robustness or camera invariance. 
Multi-view data also provides an efficient and scalable form of augmentation, delivering consistent gains with minimal additional collection cost compared to scaling single-view datasets. 
To address the scarcity of multi-view demonstrations, we propose \textbf{RoboNVS}, a geometry-aware framework that synthesizes physically consistent novel-view videos from monocular demonstrations and improves downstream policy learning. 
Future work will focus on improving generation fidelity and combining RoboNVS with additional augmentation strategies to build richer and more scalable robotic manipulation datasets.

\bibliographystyle{plainnat}
\bibliography{main}

\clearpage
\appendix
\section{Appendix Reading Guide}
\label{sec:appendix_reading_guide}

This appendix serves as an evidence pack for the main paper.
We organize it around the questions a reader may naturally ask after reading the main results.

\noindent\textbf{How is RoboNVS trained?}
We provide the detailed training corpus, dataset composition, and robot embodiments in Sec.~\ref{secC1}.
The fine-tuning setup for the video generation model is summarized in Sec.~\ref{secC2}.
The construction of self-supervised training pairs for RoboNVS is described in Sec.~\ref{secC3}.
The budget-matched sampling rules are stated in Sec.~\ref{secCBudget-Matched} of this appendix.

\noindent\textbf{How are the RoboTwin policy experiments implemented?}
We report the policy architectures and training hyperparameters for Diffusion Policy and $\pi_0$ in Sec.~\ref{secC4}.
The stochastic initialization protocol used during evaluation is detailed in Sec.~\ref{secC5}.
Task lists and the VA/VLA split are given in Sec.~\ref{secC6} of this appendix.

\noindent\textbf{How does progressively stacking multiple viewpoints affect performance?}
Beyond adding a single additional viewpoint, we also investigate progressively stacked multi-view training in the study of Sec.~\ref{sec:stacked_multiview}.

\noindent\textbf{Are the multi-view gains robust?}
We verify that the improvement is not specific to one policy architecture by repeating the experiments with ACT in Sec.~\ref{secD1}.
We further test alternative base viewpoints in Sec.~\ref{secD2} and report cross-view robustness tests in Sec.~\ref{secD3}.

\noindent\textbf{Does the benefit extend beyond RGB observations?}
We repeat the view augmentation experiment with point-cloud observations and DP3, reported in Sec.~\ref{secE}.

\noindent\textbf{Does better generation actually help policy learning?}
We compare RoboNVS with generative baselines in RoboTwin, reporting both video metrics and downstream policy success in the generator study of Sec.~\ref{secF}.

\noindent\textbf{What makes RoboNVS better as a generator?}
We isolate mask-based inpainting quality under identical masks in Sec.~\ref{secG1}.
We also compare depth prediction and alignment strategies for DW-Mesh rendering, shown in Sec.~\ref{secG2}.

\section{Supplementary Related Work}
\label{sec:related_work}

\noindent\textbf{Visual augmentation for imitation learning.}
Visual augmentation is widely used to improve the generalization of visuomotor policies trained from demonstrations.
Beyond standard image perturbations, recent works synthesize observations from new camera poses to augment robot datasets.
For example, VISTA~\cite{tian2025view} and RoVi-Aug~\cite{chen2024rovi} use image-based novel-view synthesis models such as ZeroNVS~\cite{sargent2024zeronvs}.
Other methods leverage 3D Gaussian Splatting~\cite{yang2025novel}, state-space augmentation~\cite{mitrano2022data,ke2023ccil}, or wrist-camera data augmentation~\cite{liu2025d,zhang2024diffusion}.
However, image-based view synthesis often processes frames independently, which may introduce temporal inconsistency and unrealistic robot motion.
In contrast, \hivepurple{RoboNVS} generates temporally coherent multi-view robot videos from monocular demonstrations, making the synthesized views more suitable for imitation learning.

\noindent\textbf{Video diffusion models.}
Video generation has rapidly progressed from early autoregressive or GAN-based systems to diffusion-based architectures.
Early systems such as Make-A-Video~\cite{esser2023structurecontentguidedvideosynthesis} and Gen-1~\cite{singer2022make} demonstrated text-driven video synthesis, but often struggled with long-range motion and temporal stability.
Recent models such as SVD~\cite{blattmann2023stable}, VideoCrafter~\cite{chen2023videocrafter1}, Hunyuan Video~\cite{kong2024hunyuanvideo}, CogVideoX~\cite{yang2024cogvideox}, and Wan~2.1~\cite{wang2025wan} improve temporal modeling with large-scale diffusion transformers.
Despite this progress, significant viewpoint changes remain difficult because these models primarily learn appearance priors rather than explicit scene geometry.
RoboNVS builds on the generative capacity of Wan~2.1, but introduces geometry-aware conditioning so that camera changes remain physically plausible.

\noindent\textbf{Novel view synthesis and camera control.}
Camera-controlled video generation has recently attracted increasing attention.
Some methods condition video generation on camera parameters or trajectory embeddings, such as CameraCtrl~\cite{he2024cameractrl}.
Others use intermediate 3D representations to guide camera motion and novel-view synthesis, including TrajectoryCrafter~\cite{yu2025trajectorycrafter} and CogNVS~\cite{chen2025reconstruct}.
ReCamMaster~\cite{bai2025recammaster} further enables camera control in video diffusion models, but relies on synchronized multi-camera data.
Our method is most closely related to EX-4D~\cite{hu2025ex}, which introduces Depth Watertight Mesh (DW-Mesh) as a geometry-aware prior for video generation.
However, EX-4D is designed for general dynamic camera trajectories and is trained on in-the-wild videos.
RoboNVS adapts this line of work to robotic manipulation by synthesizing fixed target-view demonstrations and adapting inpainting to the self-occlusions of a moving manipulator.

\noindent\textbf{Video depth estimation.}
Depth estimation is a key component for geometry-aware video synthesis.
One line of work focuses on temporally consistent relative depth, such as VDA~\cite{chen2025video} and DepthCrafter~\cite{hu2025depthcrafter}.
Another line predicts metric depth and camera parameters using 3D-aware foundation models, including DA3~\cite{lin2025depth}, MegaSAM~\cite{li2025megasam}, Pi3~\cite{wang2025pi}, and ViPE~\cite{huang2025vipe}.
Relative depth models are often temporally smooth but lack metric scale, while metric 3D models provide camera geometry but may contain temporal artifacts.
RoboNVS combines DepthCrafter and DA3 through inverse-depth scale-shift alignment, obtaining depth maps that are both temporally stable and geometrically grounded.

\section{Additional Implementation Details}
\label{sec:exp_details}

\subsection{Detailed Dataset Statistics}
\label{secC1}

\noindent\textbf{Training corpus.}
To train RoboNVS across diverse robotic settings, we construct a large-scale manipulation video corpus by aggregating multiple public datasets.
The corpus primarily builds upon the Open X-Embodiment (OXE)~\cite{o2024open} meta-dataset.
From OXE, we select both large-scale general robot datasets and specialized manipulation datasets.
We further incorporate additional non-OXE datasets to increase embodiment coverage and scene diversity.

\begin{table*}[t]
\centering
\caption{
Training data sources used for RoboNVS.
The corpus is primarily built from OXE subsets, with additional robotic manipulation datasets added for broader embodiment coverage.
}
\label{tab:training_datasets}
\begingroup
\small
\setlength{\tabcolsep}{5pt}
\renewcommand{\arraystretch}{1.13}
\resizebox{0.85\textwidth}{!}{%
\begin{tabular}{ll}
\toprule
\textbf{Source Family} & \textbf{Datasets} \\
\midrule
\textbf{OXE: large-scale robot datasets}
& Droid~\cite{khazatsky2024droid}, RT-1~\cite{brohan2022rt}, BridgeData V2~\cite{walke2023bridgedata} \\
\addlinespace
\textbf{OXE: specialized manipulation datasets}
& mimicplay~\cite{wang2023mimicplay}, ucsd\_kitchen~\cite{ucsd_kitchens},
ur5\_columbia\_cairlab\_pusht\_real~\cite{chi2023diffusionpolicy} \\
& utokyo\_xarm\_pick\_and\_place~\cite{matsushima2023weblab},
nyu\_franka\_play\_dataset~\cite{cui2022play}, HYDRA~\cite{belkhale2023hydra} \\
\addlinespace
\textbf{Additional robot datasets}
& RH20T~\cite{fang2023rh20t}, RoboTwin~\cite{chen2025robotwin}, RoboCoin~\cite{wu2025robocoin} \\
\bottomrule
\end{tabular}%
}
\endgroup
\end{table*}

\begin{figure*}[t]
\centering
\includegraphics[width=0.75\linewidth]{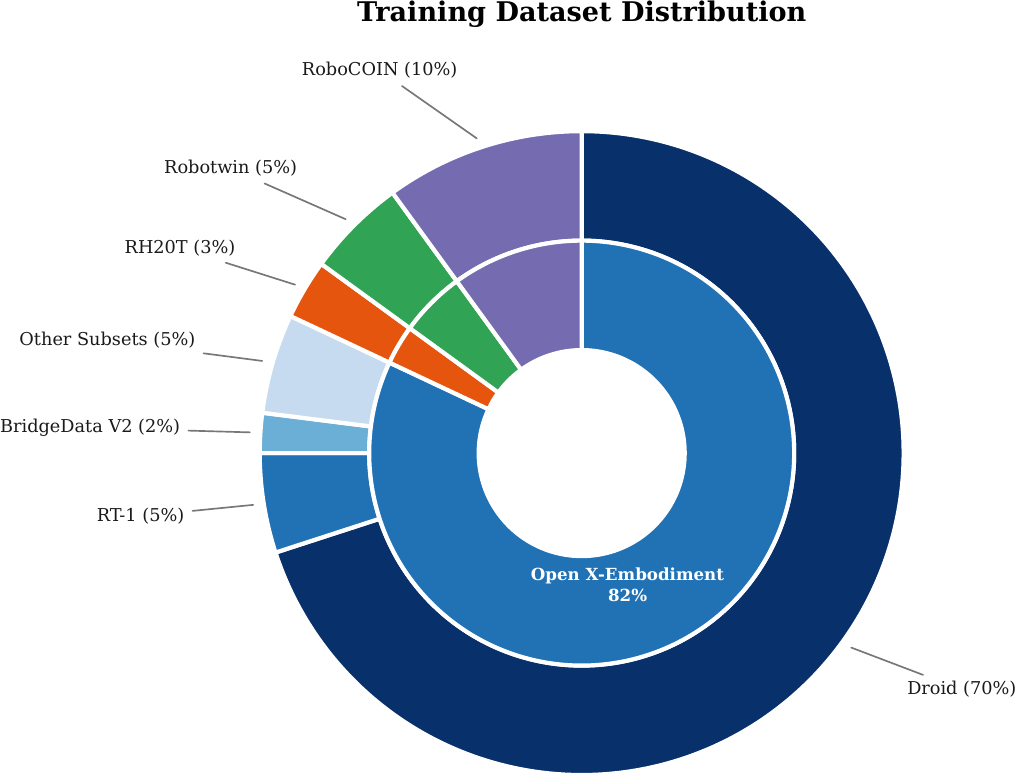}
\caption{
\textbf{Dataset composition used for training RoboNVS.}
The training corpus contains approximately 10{,}000 videos, with Droid forming the largest subset due to its scene diversity and broad manipulation coverage.
}
\label{fig:dataset_dis_pie}
\end{figure*}

\begin{table*}[t]
\centering
\caption{
Robotic embodiments and platforms covered by the aggregated training corpus.
}
\label{tab:embodiments}
\begingroup
\small
\setlength{\tabcolsep}{5pt}
\renewcommand{\arraystretch}{1.13}
\resizebox{0.88\textwidth}{!}{%
\begin{tabular}{ll}
\toprule
\textbf{Category} & \textbf{Robot Embodiments} \\
\midrule
\textbf{Standard arms}
& Franka, UR5, Sawyer, Google Robot, WidowX, Xarm \\
\addlinespace
\textbf{Mobile / dual-arm / multi-purpose platforms}
& Unitree G1 edu-u3, Mobile Aloha, A1X, AIRBOT\_MMK2, alpha\_bot\_2 \\
& AgilexCobotMagic, AgilexSplitALOHA, Galaxea R1 Lite \\
\bottomrule
\end{tabular}%
}
\endgroup
\end{table*}

In total, the dataset contains approximately 10{,}000 video sequences across diverse manipulation tasks, environments, and embodiments.
Droid constitutes approximately 70\% of the training corpus.
We use a large portion of Droid because it contains diverse real-world manipulation scenes with relatively fewer repeated demonstrations in the same environment.
This diversity helps RoboNVS learn broader visual and geometric priors for robotic manipulation.
The dataset composition is visualized in Fig.~\ref{fig:dataset_dis_pie}, and the robot embodiments are summarized in Table~\ref{tab:embodiments}.
The corpus therefore spans standard arms, mobile manipulators, dual-arm systems, and multi-purpose platforms.
This mix gives the generator a broader prior over robotic appearances before it is asked to complete a new scene.

\subsection{RoboNVS Training Details}
\label{secC2}

We fine-tune the Wan2.1-14B~\cite{wan2025wan} text-to-video foundation model using LoRA~\cite{hu2022lora} with rank $r=16$, so that only a small set of adapters is updated while the frozen backbone keeps its video prior.
All videos are resized to a spatial resolution of $320 \times 192$ with a temporal length of 49 frames.
We optimize the model using AdamW with a learning rate of $2\times10^{-5}$.
The same spatial resolution, clip length, and text prompt are reused at inference without further retuning.

The model is trained on a high-performance compute cluster equipped with \textbf{8$\times$ NVIDIA H200 (141GB)} GPUs.
To fit the 14B video model and high-resolution video latents, we use a batch size of \textbf{1} per GPU, resulting in an effective batch size of \textbf{8}.
Under this configuration, the full fine-tuning process takes approximately \textbf{one week}.
We freeze the 14B backbone and update only LoRA adapters, so robot-domain adaptation does not overwrite the pretrained motion prior.

\subsection{Self-supervised Data Generation for RoboNVS}
\label{secC3}

RoboNVS does not require synchronized multi-view supervision.
Instead, it constructs self-supervised training pairs from monocular robot videos.

\noindent\textbf{Geometry construction.}
Given a robot manipulation video $V=\{I_t\}_{t=1}^{T}$ captured by a static camera, we first estimate a depth map $D_t$ for each frame.
With camera intrinsics $K$, each pixel $\mathbf{u}=(u,v)$ is back-projected into 3D:
\begin{equation}
\mathbf{x}_t(\mathbf{u})
=
D_t(\mathbf{u})K^{-1}
\begin{bmatrix}
u & v & 1
\end{bmatrix}^{\top}.
\end{equation}
The resulting point clouds are converted into a dynamic DW-Mesh representation $\mathcal{M}_t$ following the EX-4D framework~\cite{hu2025ex}.

\noindent\textbf{Virtual target camera.}
To simulate novel-view observations, we sample a target rotation angle from two symmetric ranges:
\begin{equation}
\theta \sim \mathcal{U}\left([-60^\circ,-20^\circ]\cup[20^\circ,60^\circ]\right).
\end{equation}
Let $\mathbf{c}$ denote the depth center of the first frame.
We construct a virtual camera by rotating around $\mathbf{c}$:
\begin{equation}
G^{\theta}
=
T(\mathbf{c})R_y(\theta)T(-\mathbf{c})G^{0},
\end{equation}
where $G^{0}$ is the original camera pose, $R_y(\theta)$ is the yaw rotation, and $T(\cdot)$ denotes translation.
The target camera $G^{\theta}$ stays fixed throughout the clip, matching a static third-person camera.

\noindent\textbf{Mask construction.}
We render the mesh sequence $\{\mathcal{M}_t\}_{t=1}^{T}$ from the target camera $G^{\theta}$.
This produces a projected visibility mask $M_t^{\theta}$ on the source image grid:
\begin{equation}
M_t^{\theta}(\mathbf{u})
=
\mathbb{I}\left[
\mathbf{u}\in
\Pi_{0}
\left(
\mathcal{V}(\mathcal{M}_t,G^{\theta})
\right)
\right],
\end{equation}
where $\mathcal{V}(\mathcal{M}_t,G^{\theta})$ denotes the surface region visible under the virtual target camera, $\Pi_{0}$ projects the visible region back to the original image grid, and $\mathbb{I}[\cdot]$ is the indicator function.

The masked input frame is then constructed by preserving pixels inside the visible region and masking out the rest:
\begin{equation}
\tilde{I}_t^{\theta}(\mathbf{u})
=
M_t^{\theta}(\mathbf{u}) \odot I_t(\mathbf{u}).
\end{equation}
Equivalently,
\begin{equation}
\tilde{I}_t^{\theta}(\mathbf{u})
=
\begin{cases}
I_t(\mathbf{u}), & M_t^{\theta}(\mathbf{u})=1,\\
0, & M_t^{\theta}(\mathbf{u})=0.
\end{cases}
\end{equation}
The original unmasked frame $I_t$ serves as the reconstruction target.

\noindent\textbf{Bi-directional self-supervision.}
To increase mask diversity, we further apply the Bi-directional Masking Strategy from the main paper.
For each mask $M_t^{\theta}$, we construct its complement:
\begin{equation}
\bar{M}_t^{\theta}=1-M_t^{\theta}.
\end{equation}
This gives a complementary masked frame:
\begin{equation}
\bar{I}_t^{\theta}(\mathbf{u})
=
\bar{M}_t^{\theta}(\mathbf{u}) \odot I_t(\mathbf{u}).
\end{equation}
Thus, each monocular video yields two self-supervised training pairs:
\begin{equation}
\mathcal{P}^{\theta}
=
\left\{
(\tilde{V}^{\theta},M^{\theta},V),
(\bar{V}^{\theta},\bar{M}^{\theta},V)
\right\},
\end{equation}
where $\tilde{V}^{\theta}=\{\tilde{I}_t^{\theta}\}_{t=1}^{T}$ and $\bar{V}^{\theta}=\{\bar{I}_t^{\theta}\}_{t=1}^{T}$.
The first pair trains the model to recover regions occluded by target-view rendering, while the second pair exposes the model to complementary background and peripheral masks.
In this way, self-supervision turns monocular videos into view-change inpainting tasks: the training target remains the original video, and the input is a geometry-induced masked version of the same sequence.
RoboNVS can therefore learn target-view completion from monocular clips alone, without needing synchronized multi-view labels.

\subsection{Budget-Matched Sampling Strategies}
\label{secCBudget-Matched}

\noindent\textbf{Multi-view training experiments.}
For budget-matched multi-view experiments, we construct the multi-view training set with the same number of demonstrations as the single-view baseline. Given a single-view dataset containing $N$ trajectories, where each trajectory corresponds to a unique manipulation scene, we ensure that every original trajectory is selected exactly once during sampling. Instead of increasing the number of demonstrations, we replace the original $0^\circ$ observation with an alternative viewpoint when sampling. The viewpoints are cycled to evenly distribute viewpoint supervision across trajectories. For example, for the $\pm10^\circ$ setting with three available viewpoints ($-10^\circ$, $0^\circ$, and $10^\circ$), we cycle through:
$-10^\circ \rightarrow 0^\circ \rightarrow 10^\circ \rightarrow -10^\circ \rightarrow \cdots$,
so that each trajectory appears exactly once while being assigned different viewpoints. This strategy preserves the original scene and trajectory coverage while introducing viewpoint diversity under a fixed data budget.

\noindent\textbf{2D augmentation comparison experiments.}
For the comparison between multi-view supervision and conventional 2D visual augmentations, all augmentation-based methods are trained with the same number of demonstrations. Specifically, given 50 original $0^\circ$ demonstrations, we construct each augmented dataset by keeping 25 demonstrations unchanged and applying the corresponding augmentation method to the remaining 25 demonstrations. The resulting dataset therefore contains 25 original samples and 25 augmented samples, while maintaining the same total training budget and scene coverage as the single-view baseline. For the mixed augmentation setting, we further construct a 50-sample dataset consisting of 10 original demonstrations and four types of augmented demonstrations, with 10 samples generated from each augmentation type.
This matching keeps the comparison focused on the type of visual perturbation, not on simply adding extra demonstration videos.

The four operators use the following ranges.
Random perspective applies a distortion scale sampled from $[0.2,0.4]$.
Random rotation samples an angle from $[-30^\circ,+30^\circ]$.
Random cropping uses no padding: a window of $0.8$ of the original height and $0.8$ of the original width is placed at a random location, and the crop is then resized back to the source resolution.
Gaussian blur uses a kernel of size $5$ and $\sigma=0.5$.
The mixed setting draws from the same four operators with these ranges. 

\subsection{Policy Architectures and Training Configurations}
\label{secC4}

This section provides additional details for the policy architectures used in the RoboTwin experiments.
We evaluate two representative visuomotor policies: Diffusion Policy (DP)~\cite{chi2023diffusionpolicy} and $\pi_0$~\cite{black2024pi_0}.

For DP, we use a ResNet-18~\cite{he2016identity} visual encoder and a three-frame observation window.
The prediction horizon is set to 8, with 3 observation steps and 6 executed action steps.
For $\pi_0$, we follow the official design based on a PaliGemma-2B vision-language backbone, SigLIP visual encoder, and Gemma-300M action expert, fine-tuned with LoRA.
The key hyperparameters are summarized in Tables~\ref{tab:dp_hyperparams} and~\ref{tab:pi0_hyperparams}.
Unless a later subsection names a different architecture, the RoboTwin studies reuse these DP and $\pi_0$ settings so that training budgets stay aligned with the main text.

\begin{table}[t]
\centering
\caption{Key training hyperparameters for Diffusion Policy (DP).}
\label{tab:dp_hyperparams}
\begingroup
\small
\setlength{\tabcolsep}{4pt}
\renewcommand{\arraystretch}{1.08}
\resizebox{\linewidth}{!}{%
\begin{tabular}{l c l c}
\toprule
\multicolumn{4}{c}{\cellcolor{hiveHoney!18}\textbf{Diffusion Policy Training Setup}} \\
\midrule
Observation modality & head\_camera + agent\_pos & Image encoder & ResNet-18 \\
Image resolution & $256 \times 192$ & Batch size & 128 \\
Observation steps ($n_{\text{obs}}$) & 3 & Optimizer & AdamW \\
Prediction horizon & 8 & Learning rate & $1\times10^{-4}$ \\
Executed actions ($n_{\text{action}}$) & 6 & LR schedule & cosine decay \\
Diffusion type & DDPM & Weight decay & $1\times10^{-6}$ \\
Training diffusion steps & 100 & EMA decay & 0.99 \\
UNet channels & [256, 512, 1024] & Time embedding dim & 128 \\
\bottomrule
\end{tabular}%
}
\endgroup
\end{table}

\begin{table}[t]
\centering
\caption{Key training hyperparameters for $\pi_0$.}
\label{tab:pi0_hyperparams}
\begingroup
\small
\setlength{\tabcolsep}{4pt}
\renewcommand{\arraystretch}{1.08}
\resizebox{\linewidth}{!}{%
\begin{tabular}{l c l c}
\toprule
\multicolumn{4}{c}{\cellcolor{hiveHoney!18}\textbf{$\pi_0$ Training Setup}} \\
\midrule
Backbone VLM & PaliGemma 2B + LoRA & Precision & bfloat16 \\
Vision encoder & SigLIP & Parallelization & FSDP (4 GPUs) \\
Action expert & Gemma 300M + LoRA & Optimizer & AdamW \\
Peak learning rate & $2.5\times10^{-5}$ & Final learning rate & $2.5\times10^{-6}$ \\
LR schedule & cosine decay & Warmup steps & 1000 \\
Batch size & 32 & Gradient clipping & 1.0 \\
$\beta_1,\beta_2$ & (0.9, 0.95) & Weight decay & $1\times10^{-10}$ \\
Training steps & 5k & & \\
\bottomrule
\end{tabular}%
}
\endgroup
\end{table}

\subsection{Initialization Protocols for RoboTwin Experiments}
\label{secC5}

To prevent policies from overfitting to fixed spatial layouts, we use stochastic initialization during evaluation.
For each rollout, the manipulated object poses are randomly perturbed within a predefined range.
In the Random environment mode, tabletop textures are also resampled from a large texture pool.
This ensures that the policy cannot solve the task by memorizing a fixed trajectory or relying only on static state cues.
The same initialization ranges are shared by single-view and multi-view policies, so later comparisons isolate viewpoint supervision rather than evaluation difficulty.

\begin{figure*}[t]
\centering
\includegraphics[width=0.8\textwidth]{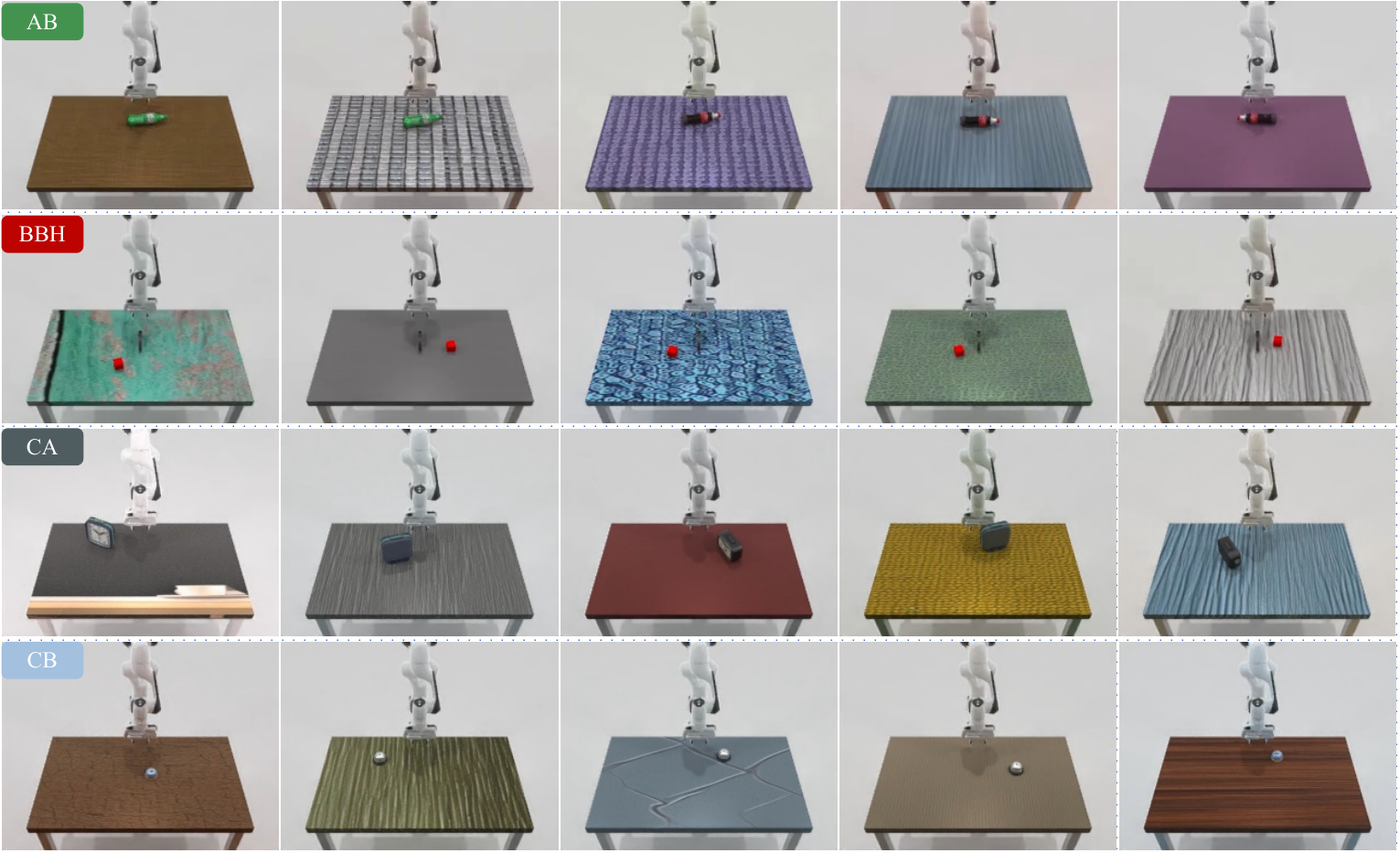}
\caption{
\textbf{Visualization of stochastic initializations.}
Each row shows one representative task, and each column shows a different randomized evaluation rollout.
The figure highlights variations in object poses and tabletop textures.
}
\label{fig:appendix_4task_firstfram}
\end{figure*}

Figure~\ref{fig:appendix_4task_firstfram} visualizes four representative tasks under five randomized evaluation rollouts.
The diversity of initial states and backgrounds confirms that evaluation is not performed under a fixed deterministic scene.
The same randomization is used for the policy tables that follow, so a higher success rate cannot be explained by an easier fixed layout.

\subsection{Detailed Specification of RoboTwin Manipulation Tasks}
\label{secC6}

We evaluate 12 manipulation tasks from RoboTwin 2.0.
The task abbreviations used in the main paper are defined below.

\noindent\textbf{Task definitions.}
\textbf{AB (Adjust Bottle):} adjust a bottle to an upright pose.
\textbf{BBH (Beat Block Hammer):} pick up a hammer and strike a block.
\textbf{CA (Click Alarm Clock):} press an alarm clock button.
\textbf{CB (Click Bell):} locate and click a service bell.
\textbf{MPA (Move Playing Card Away):} slide a playing card away.
\textbf{MCP (Move Can Pot):} place a can into a pot.
\textbf{OL (Open Laptop):} open a laptop hinge.
\textbf{PCP (Place Container Plate):} place a container on a plate.
\textbf{PS (Press Stapler):} press a stapler.
\textbf{PEC (Place Empty Cup):} place an empty cup.
\textbf{SB (Shake Bottle):} shake a bottle.
\textbf{SBH (Shake Bottle Horizontally):} shake a bottle horizontally.

\begin{figure*}[t]
\centering
\includegraphics[width=\textwidth]{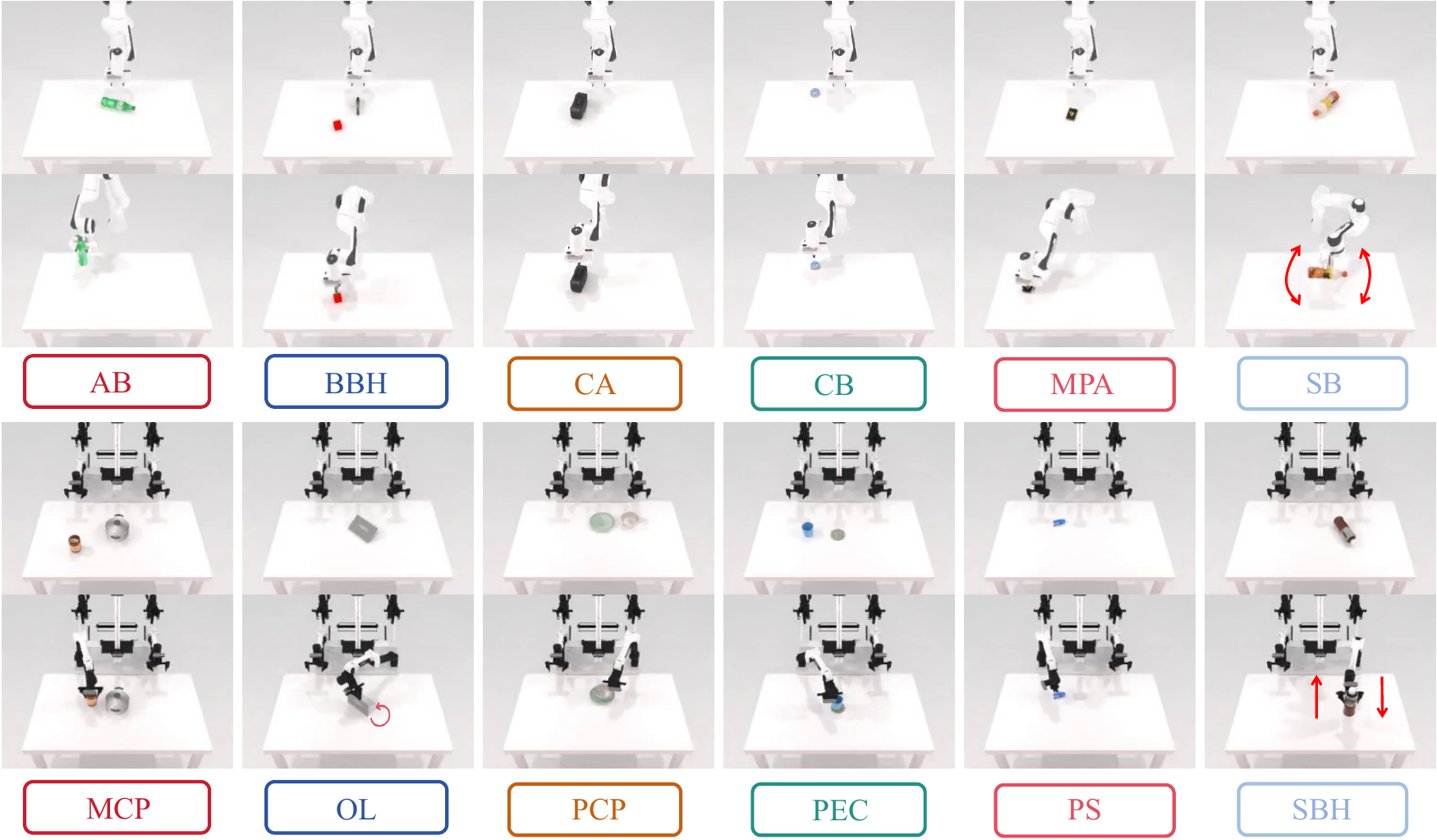}
\caption{
\textbf{Visualization of the 12 manipulation tasks in RoboTwin.}
}
\label{fig:appendix_12task_demo}
\end{figure*}

\noindent\textbf{Task categories.}
The tasks cover four representative manipulation types:
\textit{Precise Interaction} (CA, CB, PS, PEC),
\textit{Bimanual / Dynamic Manipulation} (BBH, SB, SBH),
\textit{Pick-and-Place / Relocation} (AB, MCP, PCP, MPA),
and \textit{Articulated Object Interaction} (OL).
This mix keeps the multi-view results from being tied to one manipulation family.

\noindent\textbf{Experimental task allocation.}
The VA experiment in Fig.~2 of the main paper uses six tasks:
\textit{Adjust Bottle (AB)}, \textit{Beat Block Hammer (BBH)}, \textit{Click Alarm Clock (CA)}, \textit{Click Bell (CB)}, \textit{Move Playing Card Away (MPA)}, and \textit{Shake Bottle Horizontally (SBH)}.
The VLA experiment in Table~1 uses nine seen tasks:
\textit{Adjust Bottle (AB)}, \textit{Beat Block Hammer (BBH)}, \textit{Click Alarm Clock (CA)}, \textit{Move Can Pot (MCP)}, \textit{Move Playing Card Away (MPA)}, \textit{Open Laptop (OL)}, \textit{Place Container Plate (PCP)}, \textit{Press Stapler (PS)}, and \textit{Shake Bottle Horizontally (SBH)}.
The unseen tasks are \textit{Shake Bottle (SB)}, \textit{Click Bell (CB)}, and \textit{Place Empty Cup (PEC)}.
Click Alarm Clock and Click Bell later reappear in Sec.~\ref{secF} because they require precise contact and are therefore sensitive to geometric errors in the synthesized target views.

\section{Extended Investigations on Multi-view Learning in RoboTwin}
\label{sec:robotwin_study}

\subsection{Progressive Multi-View Stacking}
\label{sec:stacked_multiview}

To further investigate whether increasing the diversity and coverage of
multi-view demonstrations continuously benefits policy learning, we conduct
an additional progressive multi-view stacking experiment. Unlike the standard
offset study in the main paper, where each setting contains the canonical
$0^\circ$ view together with only one symmetric viewpoint pair
(e.g., $0^\circ+\{\pm20^\circ\}$), here we progressively accumulate viewpoint
pairs during training. Specifically, starting from the frontal $0^\circ$ view,
we incrementally add $\pm10^\circ$, $\pm20^\circ$, $\pm30^\circ$, and so on up
to $\pm60^\circ$. Thus, the $\pm30^\circ$ stacked setting contains
$0^\circ+\{\pm10^\circ,\pm20^\circ,\pm30^\circ\}$, rather than only the
$0^\circ$ and $\pm30^\circ$ views. All policies are evaluated exclusively
under the fixed canonical $0^\circ$ viewpoint, ensuring that the experiment
measures the effect of additional multi-view supervision on the original
evaluation condition rather than viewpoint generalization.

\begin{table}[t]
\centering
\caption{
Success rates (\%) under progressively stacked multi-view training.
Starting from $0^\circ$, symmetric viewpoint pairs are incrementally added
to the training set, while all policies are evaluated at the fixed
$0^\circ$ viewpoint.
}
\label{tab:stacked_view}
\begingroup
\small
\setlength{\tabcolsep}{4.0pt}
\renewcommand{\arraystretch}{1.10}
\resizebox{\linewidth}{!}{%
\begin{tabular}{lccccc}
\toprule
\textbf{View Set} & \textbf{AB} & \textbf{BBH} & \textbf{CA} & \textbf{CB} & \textbf{Mean} \\
\midrule
$0^\circ$
& 8 & 14 & 24 & 4 & 12.5 \\

$0^\circ+\{\pm10^\circ\}$
& 54 & 30 & 46 & 18 & 37.0 \\

$0^\circ+\{\pm10^\circ,\pm20^\circ\}$
& \cellcolor{hiveHoney!12}60 & \cellcolor{hiveHoney!12}48 & \cellcolor{hiveHoney!12}58 & \cellcolor{hiveHoney!12}30 & \cellcolor{hiveHoney!12}\textbf{49.0} \\

$0^\circ+\{\pm10^\circ,\pm20^\circ,\pm30^\circ\}$
& 54 & 40 & 50 & 28 & 43.0 \\

$0^\circ+\{\pm10^\circ,\ldots,\pm40^\circ\}$
& 52 & 28 & 44 & 24 & 37.0 \\

$0^\circ+\{\pm10^\circ,\ldots,\pm50^\circ\}$
& 36 & 26 & 30 & 18 & 27.5 \\

$0^\circ+\{\pm10^\circ,\ldots,\pm60^\circ\}$
& 30 & 26 & 28 & 16 & 25.0 \\

\bottomrule
\end{tabular}%
}
\endgroup
\end{table}

\begin{table}[t]
\centering
\caption{
VLA performance under progressively stacked multi-view training.
The $\pi_0$ policy is trained with the same experimental configuration
as the main-text experiments, while all policies are evaluated at the
fixed $0^\circ$ viewpoint. (SR\%)
}
\label{tab:vla_stacked_view}
\begingroup
\small
\setlength{\tabcolsep}{4.0pt}
\renewcommand{\arraystretch}{1.10}
\resizebox{\linewidth}{!}{%
\begin{tabular}{lccccc}
\toprule
\textbf{View Set} & \textbf{AB} & \textbf{BBH} & \textbf{CA} & \textbf{CB} & \textbf{Mean} \\
\midrule

$0^\circ$
& 52 & 30 & 20 & 14 & 29.0 \\

$0^\circ+\{\pm10^\circ,\pm20^\circ,\pm30^\circ\}$
& \cellcolor{hiveHoney!12}68
& \cellcolor{hiveHoney!12}38
& \cellcolor{hiveHoney!12}38
& \cellcolor{hiveHoney!12}20
& \cellcolor{hiveHoney!12}\textbf{41.0} \\

$0^\circ+\{\pm10^\circ,\ldots,\pm60^\circ\}$
& 58 & 34 & 26 & 16 & 33.5 \\

\bottomrule
\end{tabular}%
}
\endgroup
\end{table}

The results show that progressively adding moderate viewpoint variation can
substantially improve performance under the original $0^\circ$ evaluation.
Starting from the single-view baseline with an average success rate of
$12.5\%$, adding the $\pm10^\circ$ viewpoints increases the average success
rate to $37.0\%$. The best performance is obtained when the training set
includes viewpoints up to $\pm20^\circ$, reaching an average success rate of
$49.0\%$. Further expanding the viewpoint range does not provide additional
benefits and instead leads to a gradual degradation, with the average success
rate decreasing to $43.0\%$, $37.0\%$, $27.5\%$, and $25.0\%$ when viewpoints
are expanded to $\pm30^\circ$, $\pm40^\circ$, $\pm50^\circ$, and $\pm60^\circ$,
respectively.

This trend is consistent with our standard offset study and further suggests
that the benefit of multi-view supervision is not simply determined by the
amount of additional data. Moderate viewpoint diversity provides useful
complementary visual supervision while remaining sufficiently consistent with
the canonical observation condition. In contrast, excessively large viewpoint
variations may introduce increasingly different visual appearances and
spatial configurations, which can make the learning problem more difficult.
Importantly, since every policy is evaluated at the same $0^\circ$ viewpoint,
these improvements reflect better policy learning from multi-view
demonstrations, not robustness under a shifted test viewpoint.

We further conduct the same progressive multi-view stacking experiment
with $\pi_0$ as the VLA policy under the same experimental configuration
used in the main text. The results show that the benefit of multi-view
supervision also extends to VLA models: stacking viewpoints from
$0^\circ$ to $\pm30^\circ$ substantially improves the success rate over
the single-view baseline, achieving the best performance across all four
tasks. However, further expanding the viewpoint range to $\pm60^\circ$
leads to a performance drop, consistent with the trend observed for VA.
This result further demonstrates that moderate multi-view supervision can
improve policy learning even when every policy is still tested at the
canonical $0^\circ$ view.

To further rule out the effect of increased training data,
we conduct a strict budget-matched experiment, where each
progressive multi-view setting is resampled to use the same
number of training demonstrations and training steps as the
$0^\circ$ baseline. Here, the viewpoint settings follow the
same progressive stacking protocol, i.e., $\pm10^\circ$ denotes
$0^\circ+\{\pm10^\circ\}$, while $\pm20^\circ$ denotes
$0^\circ+\{\pm10^\circ,\pm20^\circ\}$, and so on.
All policies are evaluated at the canonical $0^\circ$ viewpoint.

\begin{table}[t]
\centering
\caption{
Budget-matched progressive multi-view stacking under fixed
$0^\circ$ evaluation. (SR\%)
}
\label{tab:budget_matched_stacking}
\begingroup
\small
\setlength{\tabcolsep}{4.0pt}
\renewcommand{\arraystretch}{1.10}
\resizebox{\linewidth}{!}{%
\begin{tabular}{lccccc}
\toprule
\textbf{View Set} & \textbf{AB} & \textbf{BBH} & \textbf{CA} & \textbf{CB} & \textbf{Mean} \\
\midrule
$0^\circ$
& 8 & 14 & 24 & 4 & 12.5 \\

$0^\circ+\{\pm10^\circ\}$
& 24 & 18 & 26 & 16 & 21.0 \\

$0^\circ+\{\pm10^\circ,\pm20^\circ\}$
& \cellcolor{hiveHoney!12}34 & \cellcolor{hiveHoney!12}20 & \cellcolor{hiveHoney!12}30 & \cellcolor{hiveHoney!12}18 & \cellcolor{hiveHoney!12}\textbf{25.5} \\

$0^\circ+\{\pm10^\circ,\pm20^\circ,\pm30^\circ\}$
& 28 & 16 & 28 & 12 & 21.0 \\

$0^\circ+\{\pm10^\circ,\ldots,\pm40^\circ\}$
& 26 & 16 & 26 & 8 & 19.0 \\

$0^\circ+\{\pm10^\circ,\ldots,\pm50^\circ\}$
& 18 & 14 & 24 & 8 & 16.0 \\

$0^\circ+\{\pm10^\circ,\ldots,\pm60^\circ\}$
& 16 & 14 & 22 & 6 & 14.5 \\

\bottomrule
\end{tabular}%
}
\endgroup
\end{table}

The results remain consistent with the standard stacking study:
the $\pm20^\circ$ setting achieves the best average success rate
of $25.5\%$, outperforming the $0^\circ$ baseline of $12.5\%$ even
under a fixed data budget. This further suggests that the observed
improvement is not simply due to seeing more training videos.

\subsection{Validation on ACT}
\label{secD1}

To verify that the multi-view benefit is not specific to Diffusion Policy, we repeat the core experiment using Action Chunking with Transformers (ACT)~\cite{zhao2023learning}.
The ACT configuration is summarized in Table~\ref{tab:act_config}.

\begin{table}[t]
\centering
\caption{Key architectural and training hyperparameters for ACT.}
\label{tab:act_config}
\begingroup
\small
\setlength{\tabcolsep}{4pt}
\renewcommand{\arraystretch}{1.08}
\resizebox{\linewidth}{!}{%
\begin{tabular}{ll|ll}
\toprule
\multicolumn{4}{c}{\cellcolor{hiveHoney!18}\textbf{ACT Policy Configuration}} \\
\midrule
Backbone & ResNet-18 & Hidden dim & 512 \\
Encoder layers & 4 & Decoder layers & 7 \\
Attention heads & 8 & Feedforward dim & 3200 \\
Dropout & 0.1 & Position embedding & sine \\
Batch size & 16 & Training epochs & 6000 \\
Learning rate & $1\times10^{-5}$ & Backbone LR & $1\times10^{-5}$ \\
Weight decay & $1\times10^{-4}$ & Gradient clip & 0.1 \\
Chunk size & 50 & KL weight & 10 \\
\bottomrule
\end{tabular}%
}
\endgroup
\end{table}

We use the same demonstration data as the DP experiments.
Starting from the frontal $0^\circ$ view, we progressively construct the same view sets:
$0$,
$0+\{\pm10^\circ\}$,
$0+\{\pm20^\circ\}$,
and $0+\{\pm30^\circ\}$.
All policies are evaluated under the canonical $0^\circ$ view.
Both training and evaluation are conducted in the Random environment mode, where tabletop textures are randomly sampled to prevent shortcut learning from static backgrounds.

As shown in Table~\ref{tab:act_multiview}, ACT also benefits from multi-view demonstrations under fixed-view evaluation, so the gain is not tied to Diffusion Policy.
The best mean is again at $\pm20^\circ$ ($26.3\%$ vs.\ $15.0\%$ at $0^\circ$), matching the DP and VLA trend rather than an ACT-only artifact.

\begin{table}[t]
\centering
\caption{
\textbf{Success rate comparison of ACT policies under multi-view vs. single-view training.}
ACT policies trained with multi-view demonstrations consistently outperform the single-view baseline under fixed $0^\circ$ evaluation.
The best performance generally appears under moderate viewpoint offsets around $\pm20^\circ$.
}
\label{tab:act_multiview}
\begingroup
\small
\setlength{\tabcolsep}{4pt}
\renewcommand{\arraystretch}{1.0}
\begin{tabular}{lcccc}
\toprule
\textbf{Task}
& $0^\circ$
& $\pm10^\circ$
& \cellcolor{hiveHoney!12}$\pm20^\circ$
& $\pm30^\circ$
\\
\midrule
AB
& 28 & 48 & \cellcolor{hiveHoney!12}\textbf{50} & 34
\\
BBH
& 14 & 16 & \cellcolor{hiveHoney!12}\textbf{24} & 16
\\
CA
& 12 & \textbf{16} & \cellcolor{hiveHoney!12}\textbf{16} & 14
\\
CB
& 0 & 2 & \cellcolor{hiveHoney!12}\textbf{6} & 0
\\
SBH
& 28 & 34 & \cellcolor{hiveHoney!12}\textbf{50} & 46
\\
MPA
& 8 & 10 & \cellcolor{hiveHoney!12}\textbf{12} & 8
\\
\midrule
Mean
& 15.0 & 21.0 & \cellcolor{hiveHoney!12}\textbf{26.3} & 19.7
\\
\bottomrule
\end{tabular}
\endgroup
\end{table}

\subsection{View Augmentation with Alternative Base Viewpoints}
\label{secD2}

The main experiments use the frontal $0^\circ$ camera as the base view.
Here, we test whether the same conclusion holds when that working camera is no longer frontal.
We repeat the same augmentation protocol using $40^\circ$ and $90^\circ$ as alternative base viewpoints.
These correspond to a diagonal view and a side view of the robot, respectively.
Training and evaluation both stay at the chosen base camera; nearby views are stacked around it, as in the frontal study.

\begin{figure*}[t]
\centering
\includegraphics[width=0.85\textwidth]{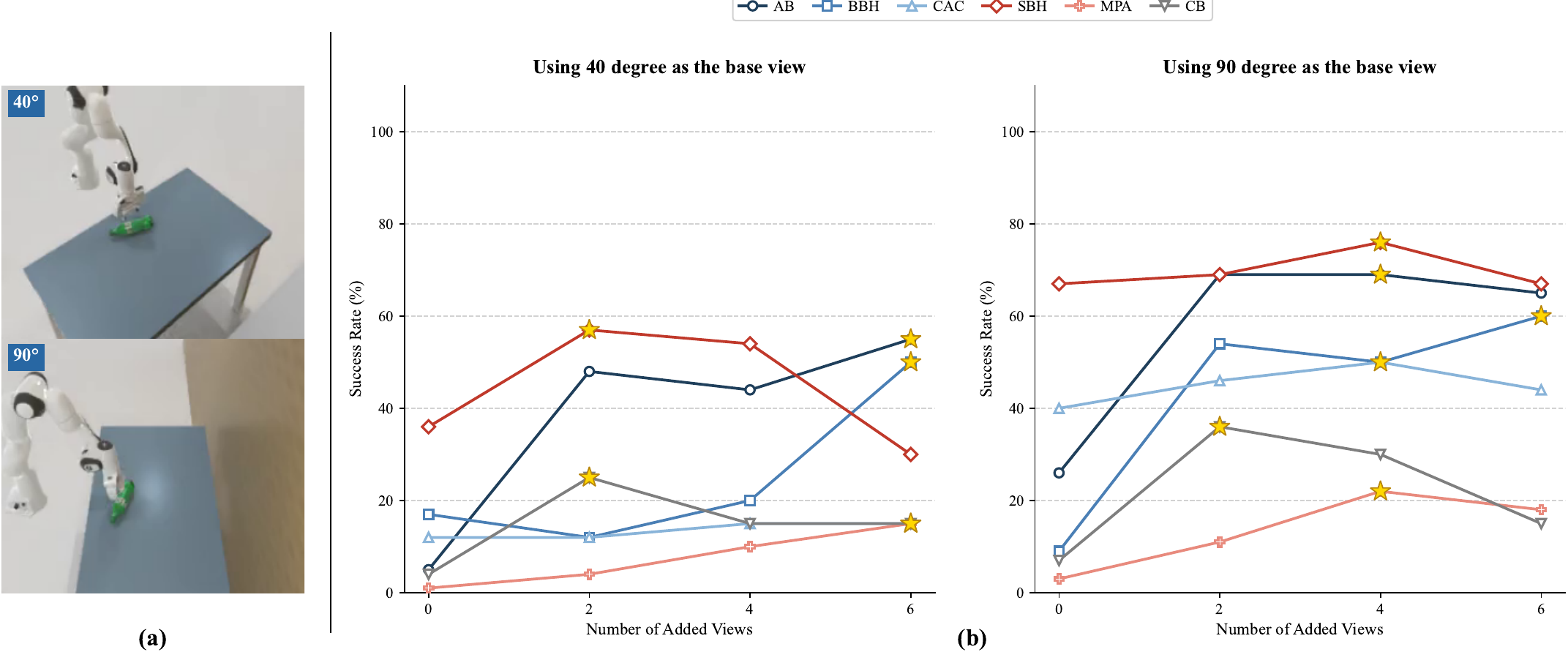}
\caption{
\textbf{View augmentation under alternative base viewpoints.}
(a) The two base viewpoints used in this study.
(b) Task success rates when the base view is $40^\circ$ or $90^\circ$.
In both settings, multi-view training improves performance over single-view training.
}
\label{fig:multi_view_40_90_with_frames}
\end{figure*}

For each base view we compare the single-view setting against three stacked ranges, adding $\pm10^\circ$, then $\pm20^\circ$, then $\pm30^\circ$ around that base.
As shown in Fig.~\ref{fig:multi_view_40_90_with_frames}, multi-view training improves success on both the $40^\circ$ and $90^\circ$ bases.
The curves again peak at a moderate stack rather than at the widest coverage.
A side camera changes how the gripper and object project into the image, yet nearby views still help more than stacking the largest offsets.

\subsection{Cross-View Robustness}
\label{secD3}

The main paper focuses on fixed-view evaluation to isolate whether multi-view data improves intrinsic manipulation capability.
Here, we evaluate the more direct benefit: robustness to camera viewpoint shifts.

We compare a single-view policy trained only on $0^\circ$ demonstrations with a multi-view policy trained on $[-20^\circ,20^\circ]$, and test both at $-20^\circ$, $-10^\circ$, $+10^\circ$, and $+20^\circ$.
Neither policy receives the test camera pose as an extra input; only which views appear in training is different.

\begin{figure*}[t]
\centering
\includegraphics[width=\textwidth]{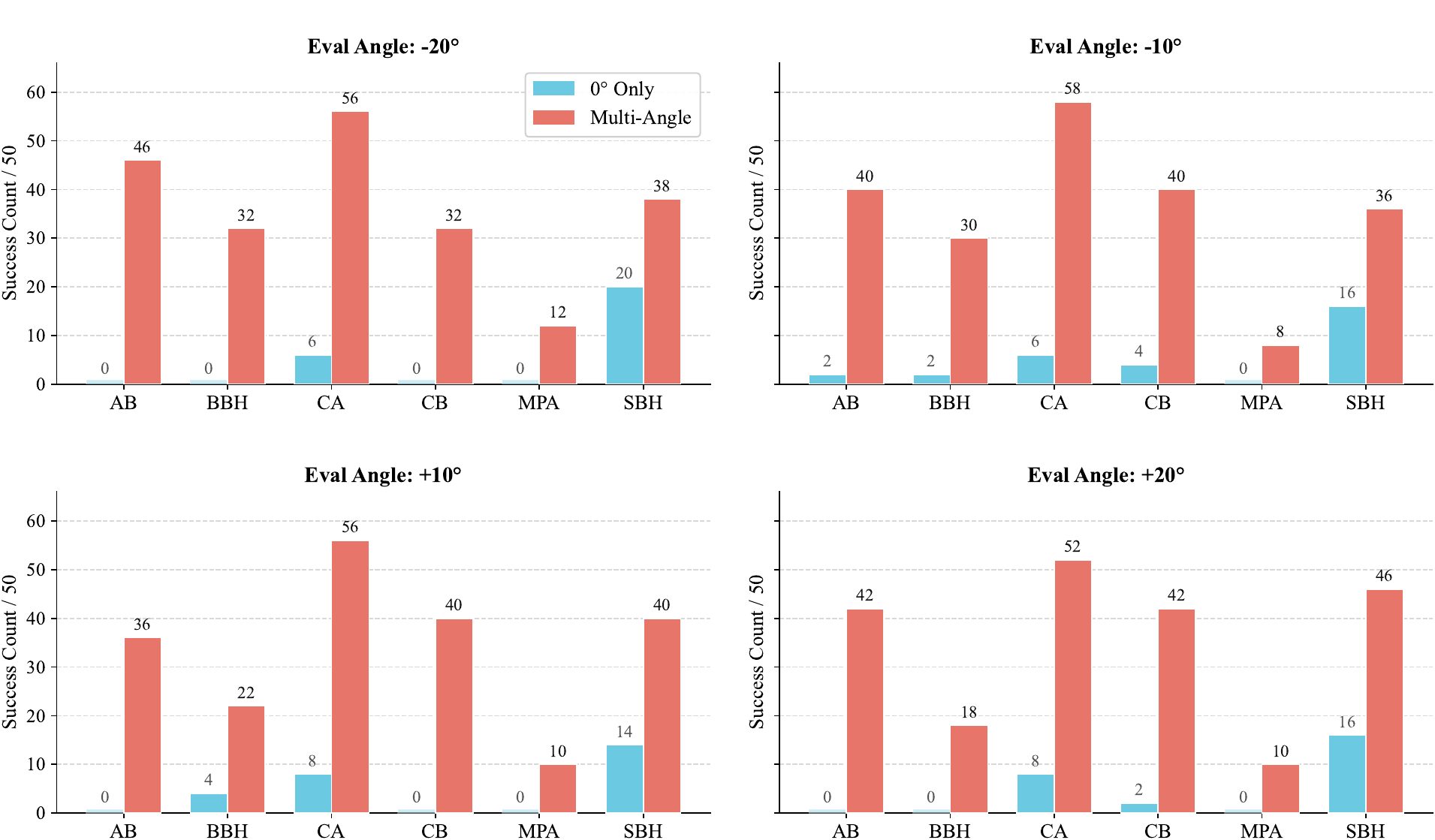}
\caption{
\textbf{Cross-view robustness evaluation.}
Policies trained with multi-view demonstrations substantially outperform single-view policies across four off-center test viewpoints.
This confirms that multi-view supervision also improves robustness when the test camera moves.
}
\label{fig:dp_bar_4panels}
\end{figure*}

Figure~\ref{fig:dp_bar_4panels} shows a consistent performance gap across tasks and viewpoints.
Single-view policies degrade sharply once the test camera leaves the training view, which is the expected cost of overfitting to a fixed appearance pattern.
The drop is especially clear on contact-rich tasks, where a small change in how the gripper and object project into the image is enough to break the visual-to-action mapping.
Multi-view policies remain substantially more reliable at all four off-center cameras.
They do not become fully view-invariant, but the success-rate gap stays large enough to be useful in deployment, where a camera is rarely placed at the exact training pose.
Although the main paper emphasizes fixed-view gains, this experiment confirms the complementary benefit: the same multi-view demonstrations that raise canonical-view success also reduce the drop when the test camera moves.

\section{View Augmentation for Point-Cloud Policies}
\label{secE}

The main VA experiments use RGB observations, so a natural concern is that the multi-view gain might be an image-space effect: extra cameras could simply diversify texture, lighting, or background statistics.
If that were the whole story, the same protocol should be much weaker when the policy no longer sees RGB frames.
To test this, we repeat the experiment with DP3~\cite{ze20243d}, replacing the visual encoder input by point clouds while keeping the rest of the comparison unchanged.

Starting from the frontal $0^\circ$ view, we construct the same progressive view sets:
$0$,
$0+\{\pm10^\circ\}$,
$0+\{\pm10^\circ,\pm20^\circ\}$,
and $0+\{\pm10^\circ,\pm20^\circ,\pm30^\circ\}$.
Demonstrations from these views are merged for training, while evaluation remains fixed at $0^\circ$.
Thus the policy still acts from a single canonical cloud at test time; the additional views are used only as training supervision.
The DP3 architecture and training setup are summarized in Table~\ref{tab:dp3_config}.
We keep the observation horizon, action horizon, and training budget aligned with the RGB Diffusion Policy experiments, so the only intended change is the observation modality.

\begin{figure*}[t]
\centering
\includegraphics[width=0.85\textwidth]{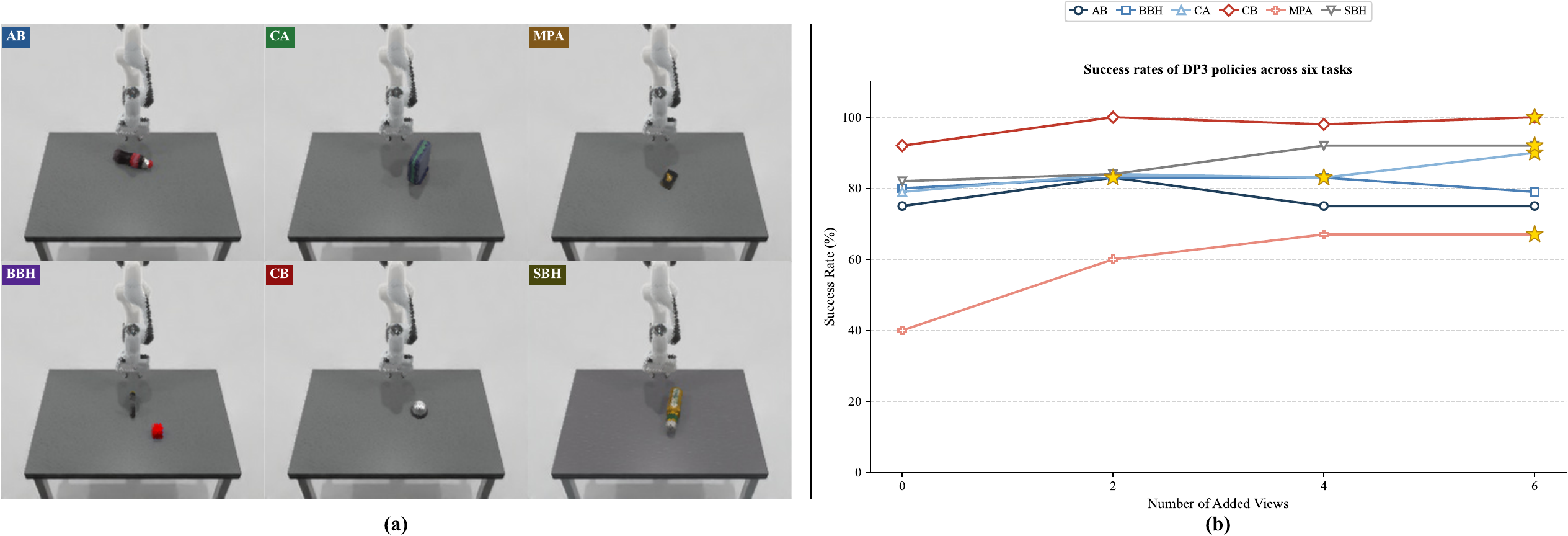}
\caption{
\textbf{View augmentation with DP3 using point-cloud observations.}
(a) Example point-cloud observations from the frontal $0^\circ$ view.
(b) DP3 success rates under progressively added viewpoints.
Multi-view training improves performance even when the policy uses 3D point clouds instead of RGB images.
}
\label{fig:dp3_linechart_new}
\end{figure*}

\begin{table*}[t]
\centering
\caption{Key architectural and training hyperparameters for DP3.}
\label{tab:dp3_config}
\begingroup
\small
\setlength{\tabcolsep}{5pt}
\renewcommand{\arraystretch}{1.10}
\resizebox{0.6\textwidth}{!}{%
\begin{tabular}{ll|ll}
\toprule
\multicolumn{2}{c|}{\cellcolor{hiveHoney!18}\textbf{Policy Architecture}}
& \multicolumn{2}{c}{\cellcolor{hiveHoney!18}\textbf{Training Setup}} \\
\midrule
Policy type & DP3 (DDIM Diffusion) & Optimizer & AdamW \\
Point cloud encoder & PointTransformer & Learning rate & $1\times10^{-4}$ \\
Transformer layers & 2 & Batch size & 16 \\
Attention heads & 2 & Training epochs & 3000 \\
Encoder output dim & 128 & Weight decay & $1\times10^{-6}$ \\
UNet channels & [512, 1024, 2048] & LR schedule & cosine \\
\midrule
\multicolumn{2}{c|}{\cellcolor{hiveHoney!12}\textbf{Observation Setup}}
& \multicolumn{2}{c}{\cellcolor{hiveHoney!12}\textbf{Action Setup}} \\
\midrule
Point cloud input & 1024 points (XYZ) & Action dim & 16 \\
Observation steps & 3 & Horizon & 8 \\
Action steps & 6 & Latency steps & 0 \\
\bottomrule
\end{tabular}%
}
\endgroup
\end{table*}

As shown in Fig.~\ref{fig:dp3_linechart_new}, multi-view training also improves DP3 across tasks.
The $0^\circ$ point-cloud baseline is already more geometric than an RGB policy, yet adding moderate nearby views still raises success.
This is consistent with the RGB study: extra viewpoints supply complementary observations of the same contact event, rather than merely more pixels of the same scene.
The gain is not uniform across tasks, but the direction is stable, and the best stacked setting again lies in a moderate range rather than at the largest coverage.
The benefit of view augmentation therefore comes from richer viewpoint coverage, not from an RGB-only observation modality.
In other words, even a 3D policy can underuse the workspace geometry if it is trained from a single camera frustum.

\section{Generative Models as Multi-View Data Engines}
\label{secF}

This section evaluates different generative models as data augmentation tools in RoboTwin.
The key question is not only which model generates better images, but which model produces synthetic views that improve downstream policy learning.
A generator can look sharp while still inventing a second gripper, deleting the target object, or breaking the arm kinematics; those errors become false demonstrations once they enter the training set.

\noindent\textbf{Experimental setup.}
We use two representative tasks: \textit{Click Alarm Clock (CA)} and \textit{Click Bell (CB)}.
Both require precise contact with a small object, so geometric mistakes in the synthesized view are likely to show up in policy success.
Each task contains 50 expert demonstrations captured from the $0^\circ$ viewpoint.
We generate four synthetic views at $-20^\circ$, $-10^\circ$, $10^\circ$, and $20^\circ$ using RoboNVS, EX-4D, TrajectoryCrafter, CogNVS, and ZeroNVS-ft.
These offsets sit in the moderate range identified by the main-text angle study, so a faithful generator should produce useful supervision rather than extreme appearance shifts.
ZeroNVS is fine-tuned on MimicGen~\cite{mandlekar2023mimicgen}, following VISTA~\cite{tian2025view}, for a fair robotic comparison.
The generated views are combined with the original $0^\circ$ data to train policies, which are then evaluated at the fixed $0^\circ$ viewpoint.
No generated frame is used at test time; the synthetic videos serve only as extra training demonstrations.

For generation-quality evaluation, each model synthesizes the same four target views and is compared against simulator ground-truth videos.
We report PSNR, SSIM, LPIPS~\cite{zhang2018unreasonable}, FID~\cite{heusel2017gans}, and FVD~\cite{heusel2018ganstrainedtimescaleupdate}.
PSNR and SSIM measure low-level reconstruction, LPIPS and FID capture perceptual fidelity, and FVD is more sensitive to temporal drift as the robot moves through the scene.

\begin{figure*}[t]
\centering
\includegraphics[width=\textwidth]{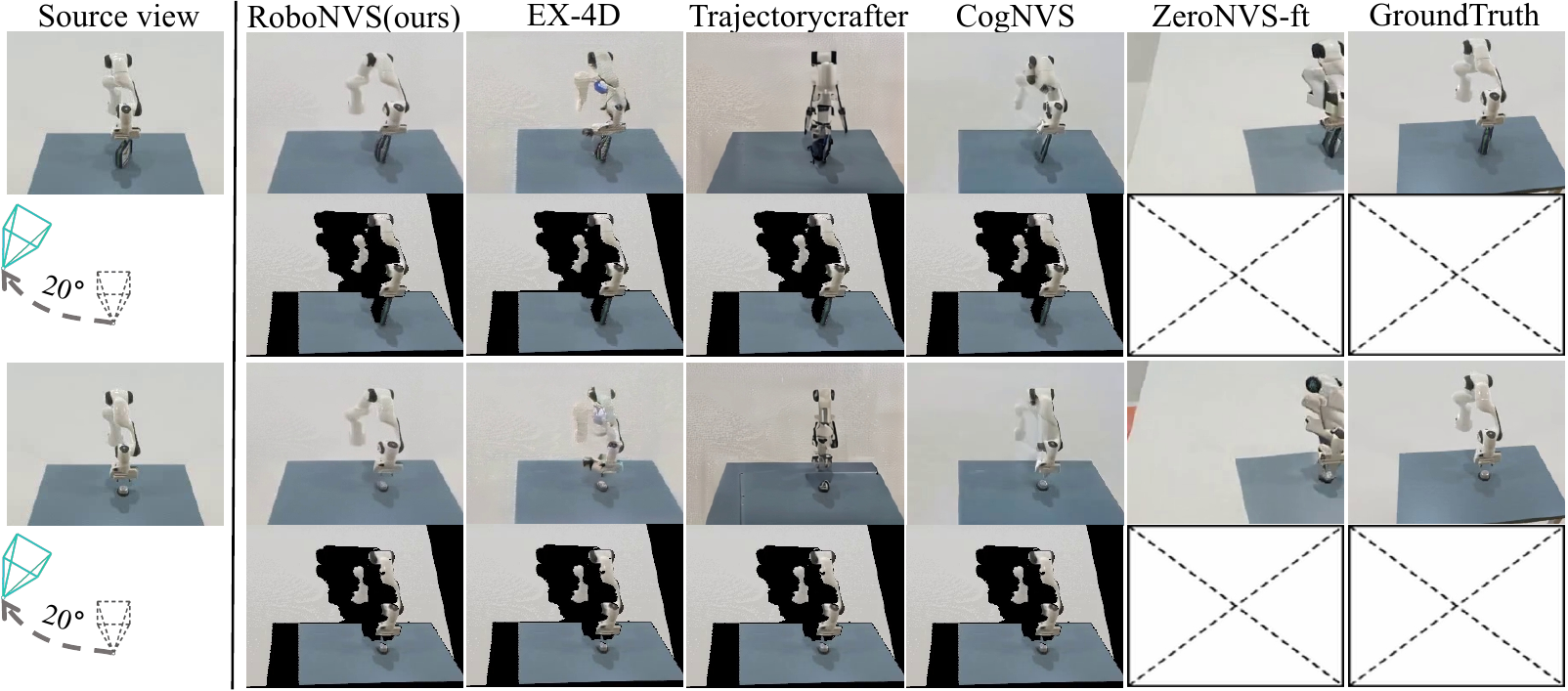}
\caption{
\textbf{Qualitative comparison of generative models under a $+20^\circ$ rotation in RoboTwin.}
All mask-based methods are conditioned on identical masks derived from simulator ground-truth depth.
RoboNVS reconstructs robot joints and object geometry more faithfully than the compared generators.
}
\label{fig:robotwin_generation_compare}
\end{figure*}

\noindent\textbf{Qualitative analysis.}
Figure~\ref{fig:robotwin_generation_compare} compares synthesized frames under a $+20^\circ$ target view.
For mask-based methods, we use ground-truth simulator depth to construct identical DW-Mesh masks, isolating the inpainting capability of each generator.
This is a favorable setting for the baselines: they are not punished for depth errors, only for how they complete the newly exposed regions.
Even then, EX-4D and CogNVS often exhibit distorted joints, blurred gripper textures, or robot links that fail to connect through the occluded region.
TrajectoryCrafter preserves some scene layout but tends to leave incomplete fills around the arm and the manipulated object.
ZeroNVS-ft, which has no explicit geometric mask, is more likely to invent inconsistent structures when the camera rotates away from the source view.
RoboNVS produces cleaner robot structures and more stable object layouts.
The completed frames keep a single coherent arm, a localized contact object, and a background that does not overwrite the workspace.
These qualitative differences matter because a policy trained on a broken arm or a missing bell is being taught the wrong contact state.

\begin{table}[t]
\centering
\caption{Quantitative comparison of generative video synthesis quality in RoboTwin.}
\label{tab:gen_quality_metrics}
\begingroup
\small
\setlength{\tabcolsep}{4pt}
\renewcommand{\arraystretch}{1.05}
\resizebox{\linewidth}{!}{%
\begin{tabular}{lccccc}
\toprule
\textbf{Model}
& \textbf{PSNR $\uparrow$}
& \textbf{SSIM $\uparrow$}
& \textbf{LPIPS $\downarrow$}
& \textbf{FID $\downarrow$}
& \textbf{FVD $\downarrow$} \\
\midrule
CogNVS & 17.154 & 0.7169 & 0.3621 & 245.13 & 47.36 \\
EX-4D & 17.031 & 0.6703 & 0.4331 & 248.26 & 79.26 \\
TrajectoryCrafter & 16.480 & 0.6687 & 0.4714 & 217.33 & 68.47 \\
ZeroNVS-ft & 14.350 & 0.7065 & 0.5465 & 215.58 & 86.18 \\
\midrule
\rowcolor{hiveSoftMint}
\textbf{RoboNVS}
& \textbf{18.839}
& \textbf{0.7575}
& \textbf{0.3018}
& \textbf{132.87}
& \textbf{33.28} \\
\bottomrule
\end{tabular}%
}
\endgroup
\end{table}

\begin{table}[t]
\centering
\caption{
Policy success rates after multi-view data augmentation using different generative models.
All policies are evaluated at the RoboTwin $0^\circ$ viewpoint over 50 trials.
}
\label{tab:gen_success_rates}
\begingroup
\small
\setlength{\tabcolsep}{5pt}
\renewcommand{\arraystretch}{1.08}
\resizebox{\linewidth}{!}{%
\begin{tabular}{lcc}
\toprule
\textbf{Training Data}
& \textbf{Click Alarm Clock}
& \textbf{Click Bell} \\
\midrule
Single-view baseline ($0^\circ$ only)
& 19/50 (38\%) & 5/50 (10\%) \\
EX-4D + $0^\circ$
& 36/50 (72\%) & 9/50 (18\%) \\
TrajectoryCrafter + $0^\circ$
& 28/50 (56\%) & 9/50 (18\%) \\
CogNVS + $0^\circ$
& 34/50 (68\%) & 35/50 (70\%) \\
ZeroNVS-ft + $0^\circ$
& 22/50 (44\%) & 6/50 (12\%) \\
\midrule
\rowcolor{hiveSoftMint}
\textbf{RoboNVS + $0^\circ$}
& \textbf{40/50 (80\%)}
& \textbf{38/50 (76\%)} \\
\bottomrule
\end{tabular}%
}
\endgroup
\end{table}

Table~\ref{tab:gen_quality_metrics} shows that RoboNVS achieves the best score on every reported metric.
The margin is largest on FID and FVD, which better reflect whether the completed video remains a plausible robot scene over time rather than a stack of individually acceptable frames.
Baselines can be close on SSIM while still producing temporally unstable arms, and those instabilities show up as a much higher FVD.
Table~\ref{tab:gen_success_rates} then asks the more important question for this paper: do the synthesized views help a policy?
The monocular baseline reaches $38\%$ on Click Alarm Clock and only $10\%$ on Click Bell.
Adding EX-4D or TrajectoryCrafter views helps the first task but barely moves the second, suggesting that their residual geometric errors are especially harmful when the target is small.
CogNVS is stronger on Click Bell, yet still trails RoboNVS on both tasks.
RoboNVS reaches $80\%$ and $76\%$, respectively.
The ranking of generators is therefore not identical to the ranking of pixel metrics, but the best generator is also the best data engine.
Better novel-view videos produce stronger policies because they preserve the gripper--object relation that a visuomotor policy has to imitate.

\section{Extended Visualizations and Comparisons}
\label{sec:visualizations}

\subsection{Mask-Based Inpainting under Identical Masks}
\label{secG1}

To isolate pure inpainting capability, we compare RoboNVS with EX-4D, TrajectoryCrafter, and CogNVS under identical mask inputs.
This removes differences caused by depth or mask construction and directly evaluates how well each model completes robotic manipulation scenes.
In the main paper, a weaker generator may fail either because its depth prior is wrong or because its video model cannot fill the resulting holes.
The comparison here freezes the first factor.
All methods receive the same mask sequences, the same number of frames, and the same source appearance reference, so remaining differences come from the completion model itself.
We focus on the failure modes that most affect later policy training, including truncated robot arms, ghosted joints, missing end-effectors, and objects that disappear or drift.
These artifacts are more damaging than mild texture blur, because they change the apparent state of the robot and the object being manipulated.

\begin{figure*}[t]
\centering
\includegraphics[width=0.82\textwidth]{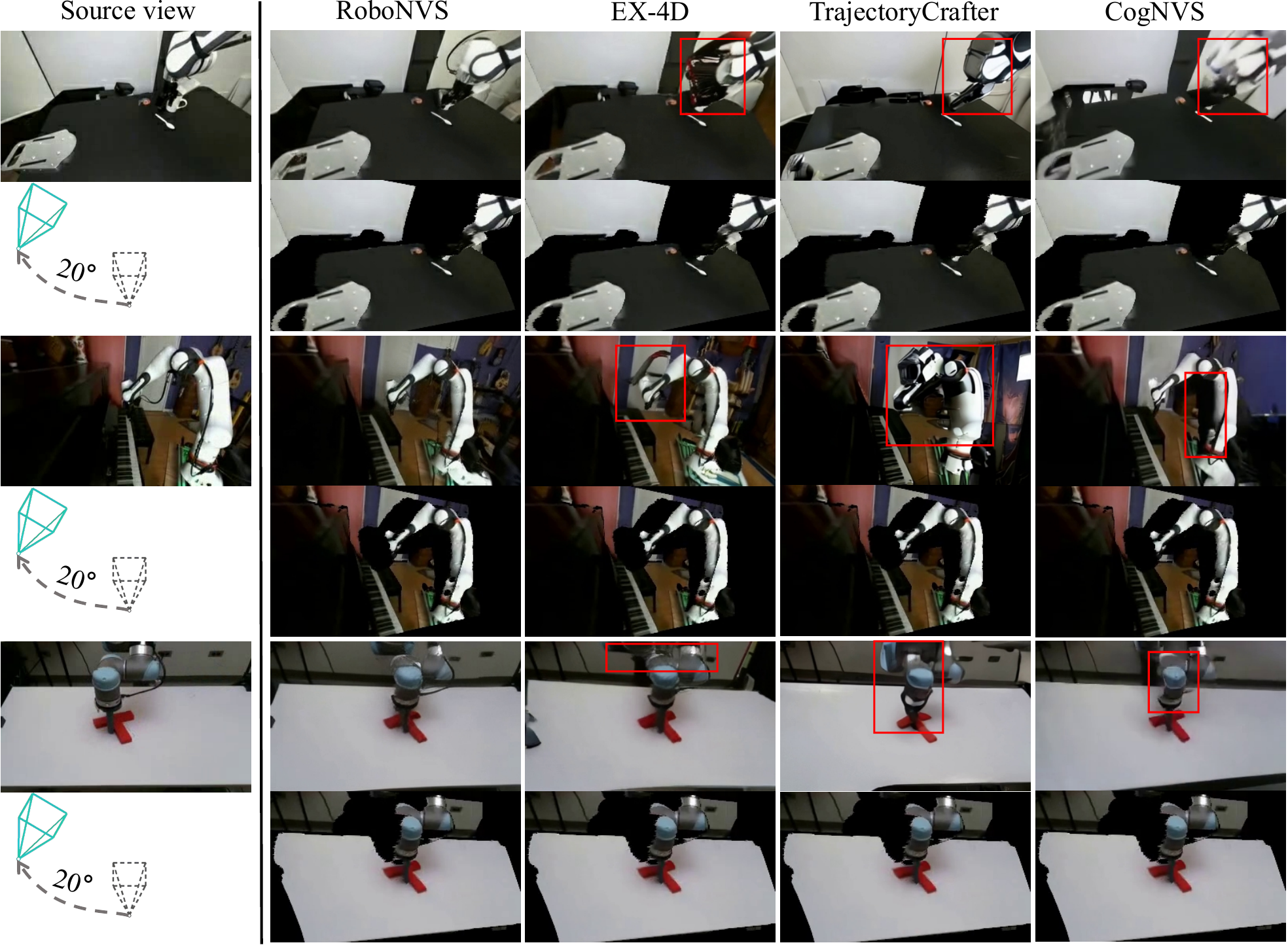}
\caption{
\textbf{Qualitative comparison of mask-based generative inpainting.}
All models receive the same mask sequences.
RoboNVS produces more complete robot structures, fewer ghosting artifacts, and fewer geometric distortions around the arm and the contacted object.
}
\label{fig:appendix_same_mask_impaint}
\end{figure*}

As shown in Fig.~\ref{fig:appendix_same_mask_impaint}, existing baselines often produce inaccurate completions, including distorted robot joints, ghosting artifacts, and missing masked regions.
The errors concentrate on the arm and the manipulated object, which are exactly the regions a visuomotor policy must attend to.
A completed frame can look globally plausible while the gripper is split, shortened, or copied, and that local error is enough to corrupt a demonstration.
We attribute these failures to two factors.
First, these models are trained on limited robotic manipulation data, so they have a weaker prior over articulated metal links and thin tools.
Second, they are primarily optimized for general dynamic camera movement, whose masks often cover background or newly revealed walls rather than a self-occluding robot arm.
In contrast, RoboNVS reconstructs sharper and more complete robotic structures, benefiting from robot-domain training and Bi-directional Masking.
The complementary mask forces the model to complete both the occluded foreground and the surrounding workspace, which reduces the hole-filling patterns that leave an arm unfinished.
Even with identical masks, it produces more faithful manipulation scenes than general geometry-aware video generators.
The remaining gap is therefore a completion-model gap, not a difference in the input mask.

\subsection{DW-Mesh Renderings from Different Depth Models}
\label{secG2}

Depth quality directly determines the quality of DW-Mesh rendering and the masks used for video completion.
RoboNVS never asks the diffusion model to invent the whole scene; it only completes what the mesh cannot explain.
If the mesh is wrong, the mask is wrong, and the generator is asked to fill the wrong region.
A noisy depth map produces broken robot meshes, incorrect occlusion boundaries, and unstable target-view masks, all of which later leak into the generated videos.
We therefore compare seven depth strategies on the same source clips:
two temporally consistent relative-depth estimators, DepthCrafter and VDA;
three metric 3D-aware models, Pi3, ViPE, and DA3;
and two alignment strategies, VDA w/ DA3 and DepthCrafter w/ DA3.
The comparison is qualitative but systematic: we inspect object scale, robot-link continuity, boundary spikes, and temporal flicker across three scenes that use different objects.

\begin{figure*}[t]
\centering
\includegraphics[width=0.78\textwidth]{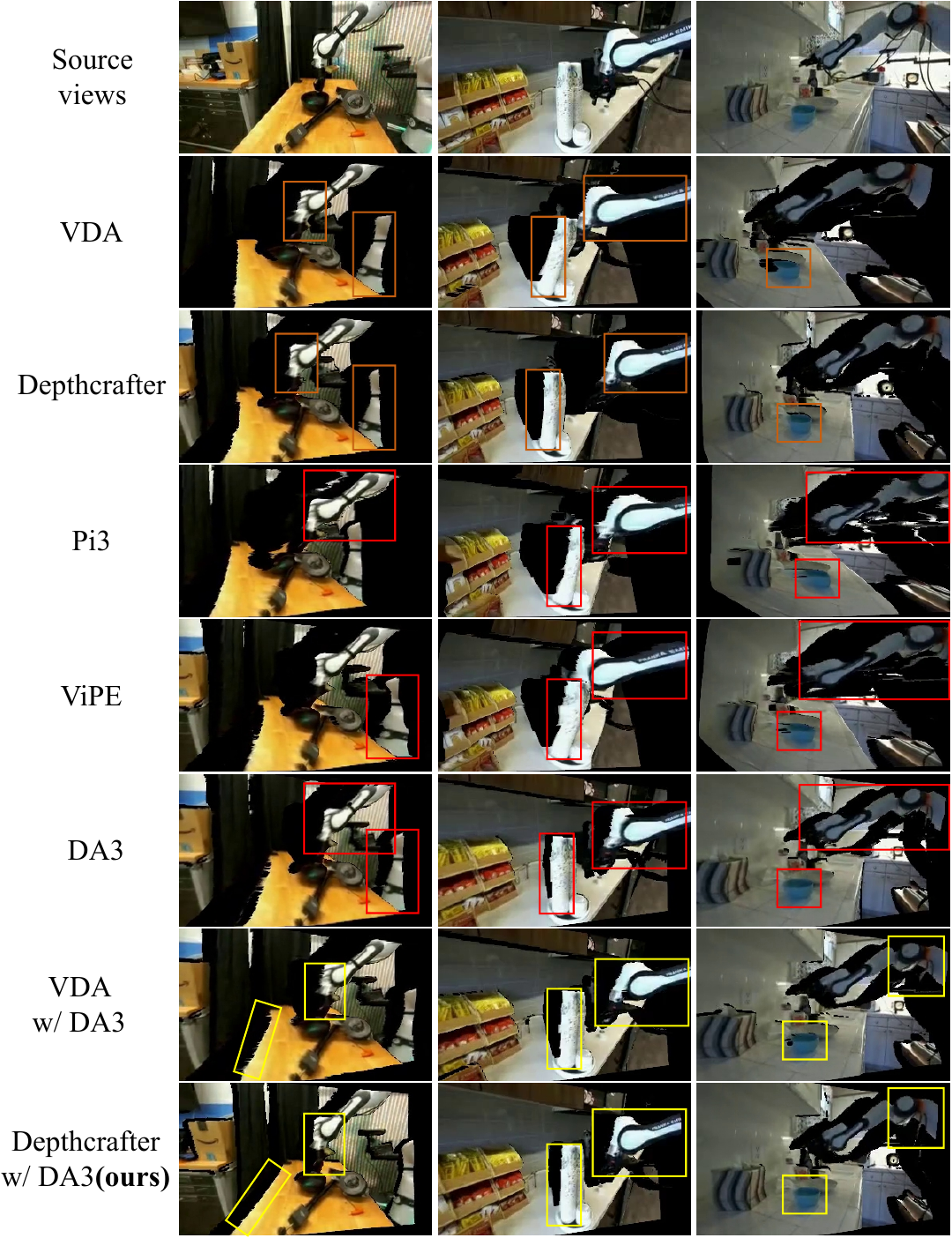}
\caption{
\textbf{Qualitative comparison of DW-Mesh renderings constructed from different video depth prediction methods.}
The first row shows source views from three manipulation scenes.
Each column visualizes DW-Mesh rendering from a different depth strategy.
Brown boxes compare temporal depth estimators, red boxes compare metric 3D-aware models, and yellow boxes highlight why DepthCrafter w/ DA3 is cleaner than the other aligned variant.
}
\label{fig:depth_mesh_compare}
\end{figure*}

\noindent\textbf{Temporal depth comparison.}
Comparing DepthCrafter and VDA, DepthCrafter produces more stable and geometrically plausible DW-Mesh renderings.
As highlighted by the brown boxes in Fig.~\ref{fig:depth_mesh_compare}, DepthCrafter yields more reasonable object scales and fewer artifacts around depth discontinuities.
Edges of the table, bottle, and robot forearm remain connected instead of breaking into isolated sheets.
VDA, in contrast, often produces distorted object shapes and boundary spikes.
These spikes are particularly harmful for novel-view rendering, because a thin depth error at an object silhouette becomes a large hole or a floating fragment once the camera rotates.
Since DW-Mesh geometry is directly determined by predicted depth, this is why RoboNVS uses DepthCrafter when predicting temporally consistent relative depth.

\noindent\textbf{Metric depth comparison.}
Relative depth is not enough for camera control: the virtual target pose needs metric scale.
Among Pi3, ViPE, and DA3, DA3 produces the cleanest metric geometry.
As highlighted by the red boxes, ViPE shows severe scale distortion, stretching the workspace so that the robot and the object no longer occupy a consistent 3D volume.
Pi3 is closer in scale but contains noticeable surface artifacts and irregular mesh structures, especially around thin links and small buttons.
DA3 gives more consistent object scales and cleaner surfaces, especially around small objects and robot structures.
Therefore, we use DA3 as the metric model that supplies camera geometry for the inverse-depth scale-shift alignment that follows.

\noindent\textbf{Depth alignment comparison.}
Neither family is sufficient alone.
Temporal models lack metric scale, while metric models can flicker or lose fine structure.
The yellow boxes compare VDA w/ DA3 and DepthCrafter w/ DA3, which attach a global scale-shift in inverse depth to a temporal estimator.
Both aligned variants improve over the raw metric models, but they are not equal.
DepthCrafter w/ DA3 yields cleaner object boundaries, fewer artifacts, and more accurate scale.
VDA w/ DA3 still inherits some of VDA's silhouette spikes, which alignment cannot fully remove.
This result shows that combining temporally stable relative depth with metric geometry is more reliable than using either component alone, and that the choice of temporal estimator still matters after alignment.
DepthCrafter contributes temporal stability and fine structural details, while DA3 provides metric scale and camera geometry; their alignment therefore yields the cleanest DW-Mesh masks for geometry-guided video completion.
These masks are the geometric prior used throughout RoboNVS, so the depth comparison is not a side study: it determines how much the diffusion model later has to invent.
Replacing any of the three components either breaks metric scale or reintroduces silhouette artifacts that later become holes in the target-view mask.

\end{document}